\documentclass{ametsocV6.1_arxiv}

\usepackage{subcaption}
\usepackage[hidelinks]{hyperref}
\newcommand\R{{\mathbb R}}
\newcommand\Z{{\mathbb Z}}

\title{A Primer on Topological Data Analysis to Support Image Analysis Tasks in Environmental Science}
\authors{
    Lander Ver Hoef,\aff{a}\correspondingauthor{Lander Ver Hoef, lander.verhoef@gmail.com}
    Henry Adams,\aff{a}
    Emily J.\ King,\aff{a}
    Imme Ebert-Uphoff\aff{b,c}
}

\affiliation{
    \aff{a}{Department of Mathematics, Colorado State University, Fort Collins, CO}\\
    \aff{b}{Dept.\ of Electrical and Computer Engineering, Colorado State University, Fort Collins, CO}\\
    \aff{c}{Cooperative Institute for Research in the Atmosphere (CIRA),
    Colorado State University, Fort Collins, CO}
}

\begin{document}

\nolinenumbers


\maketitle

%
%
%
\statement
Information such as the geometric structure and texture of image data can greatly support the inference of the physical state of an observed earth system, for example in remote sensing to determine whether wildfires are active or to identify local climate zones. 
Persistent homology is a branch of topological data analysis that allows one to extract such information in an interpretable way – unlike black box methods like deep neural networks. 
The purpose of this paper is to explain in an intuitive manner what persistent homology is and how researchers in environmental science can use it to create interpretable models.  
We demonstrate the approach to identify certain cloud patterns from satellite imagery and find that the resulting model is indeed interpretable.
%

%


\section{Introduction} 
\label{sec:intro}

Methods for image analysis have become an essential tool for many environmental science (ES) applications to automatically extract key information from satellite imagery or from gridded model output
\citep{schultz2021can,zhu2017deep,gagne2019interpretable,ebert2020evaluation}.
Machine learning (ML) methods such as convolutional neural networks (CNNs) are now the dominant technique for many such tasks, where they operate as black boxes \citep{mcgovern2019making}.
This is undesirable for high stakes applications \citep{rudin2019stop,mcgovern2022we}.
In this paper, we show how a tool that is beginning to be used in the community, namely \emph{Topological Data Analysis (TDA)}, can be combined with ML methods for interpretable image analysis.
TDA is a mathematical discipline that can quantify geometric information from an image in a predictable and well-understood way.
In Section \ref{sec:case_study}\ref{subsec:interpreting}, we give a novel example of how we can leverage this understanding to give a strong interpretation of ML results in terms of image features.

TDA has proven highly successful to aid in the analysis of data in a variety of applications, including 
neuroscience~\citep{chung2009persistence,Gardner2022Nature}, 
fluid dynamics~\citep{kramar2016analysis}, 
and cancer histology~\citep{lawson2019persistent}.
In environmental science, TDA has recently shown potential to 
help identify atmospheric rivers~\citep{muszynski2019topological},
detect solar flares~\citep{deshmukh2022machine, sun2021improved},
identify which wildfires are active~\citep{kim2019deciphering},
quantify the diurnal cycle in hurricanes~\citep{tymochko2020using}, 
identify local climate zones~\citep{sena2021topological},
detect and visualize Rossby waves~\citep{merritt2021visualizing},
and forecast COVID-19 spread using atmospheric data~\citep{segoviadominguez2021tlifelstm}.
The purpose of this article is to provide an intuitive introduction to TDA for the ES community -- using a meteorological application as guiding example -- and an understanding of where TDA might be applied.
This article is accompanied by easy-to-follow sample code provided in a GitHub repository (\url{https://github.com/zyjux/sffg_tda}) that we hope will be used by the community in new applications.

\subsection{Guiding application - analysing the mesoscale organization of clouds}
\label{subsec:guiding_app}

In order to provide a gentle introduction to TDA for the ES community, we illustrate its use for a practical example.
We chose the application of classifying the mesoscale organization of clouds, specifically distinguishing four types of organization--{\it sugar}, {\it gravel}, {\it fish}, and {\it flowers}--identified by \citet{stevens2020sugar}.
This task provides an ideal case study for our exploration of topological data analysis for several reasons: 
(1) These four organization patterns are well known from the seminal paper, \citet{stevens2020sugar}, and meteorological experts were able to reliably identify these patterns from satellite visible imagery.
(2) The task can be formulated as classification of patches of single-image monochromatic imagery, which a common TDA algorithm (\emph{persistent homology}) is well-suited for.
(3) TDA has never been applied for this application, so it is novel.
(4) A well developed benchmark data set with reliable crowd sourced labels is publicly available for this task~\citep{rasp2020combining}.

Several ML approaches have already been developed with good success for this benchmark data set to classify the four different types \citep{rasp2020combining}.
We emphasize that we are {\it not} seeking to match or exceed the performance of those ML approaches.
Rather, we use this application to demonstrate TDA as an approach that can help increase transparency, decrease computational effort, and be feasible even if few labeled data samples are available; see Section \ref{sec:intro}\ref{subsec:TDA_advantages}.

\subsection{Key TDA concepts discussed here}
\label{subsec:key_tda}

We focus on the TDA concept that is most appropriate for image analysis, {\it persistent homology}.
We provide a short preview to persistent homology below; a detailed introduction is in Section \ref{sec:tda}.

{\it Homology} is the classical study of connectivity and the presence of holes of various dimensions, giving large-scale geometric information.
{\it Persistent homology} provides a descriptor with information on the texture of an image (how rough or smooth it is) and which can be vectorized into a format useful for machine learning.
It does this by scaling through all the intensity values in an image and recording at what intensities connected components and holes appear and disappear -- see Figure \ref{fig:sub_hom_ex}.
Particularly on images, persistent homology and its vectorizations can be efficiently computed, so for image analysis (from models or satellites), the computational effort of implementing persistent homology is small.

The results of persistent homology computations can be displayed as either \emph{persistence diagrams} or \emph{persistence barcodes}.
We focus here on barcodes, in which each feature (connected component or hole) appears as a bar which starts at the intensity value at which the feature appears, and ends at the intensity at which it disappears.
The lengths of these bars indicate the \emph{persistence} of each feature.
The raw output of persistent homology is not suitable for most machine learning tasks, as the output vector varies in length from sample to sample.
While there are many proposed solutions to this, in this paper we use \emph{persistence landscapes}, which translate a barcode into a mountain range, with the height of each mountain representing the persistence of the corresponding feature -- see Figure \ref{fig:landscape_ex}.
The landscape is obtained from this mountain range by taking the $n$ highest profiles as piecewise-linear functions, where $n$ is a hyperparameter.

\subsection{Advantages of TDA for image analysis tasks in environmental science}
\label{subsec:TDA_advantages}

Persistent homology is a deterministic mathematical transformation (just as, say, the well known Fourier transform).
We first explore the advantages that persistent homology inherits from being a deterministic algorithm.
\begin{enumerate}
    \item 
       {\bf Transparency:} 
       All the internal steps of the algorithm are known and well-understood, and the method has a high degree of theoretical interpretation, giving it far more transparency than most ML methods.
       In Section~\ref{sec:case_study}\ref{subsec:interpreting}, we use this theoretical background to understand what image features are driving differences in the output of persistent homology.

    \item
       {\bf Known failure modes:} 
       No technique is perfect, and there will always be situations that cause errors and incorrect results.
       To use a method in practice, it is important to understand in what situations it struggles and what sorts of errors can result.
       Because persistent homology is a deterministic method, we can both theoretically predict these failure modes and interpret experimental results in terms of the original feature space.
       
    \item
       {\bf No need for large labeled datasets:}
       As a deterministic algorithm, persistent homology does not require large, reliably labeled datasets.
       Instead, a small set of representative examples can be used to explore the different patterns that emerge in the transformed data.
       TDA is often used in combination with a simple ML model, and the number of labeled samples to obtain good performance is smaller than would be required to train a CNN or similar tool without TDA.
       This is a huge advantage for environmental science  datasets, which are frequently large and detailed but almost entirely unlabeled.

    \item
       {\bf Environmentally friendly:}
       Many CNNs for image analysis tasks are known to have a surprisingly high carbon footprint due to the extensive computational resources required for model training \citep{schwartz2020green,xu2021survey}.
       TDA is more in line with the Green AI movement \citep{schwartz2020green,xu2021survey}, as it enables context-driven numerical results without the environmental impact inherent in training a deep neural network.

\end{enumerate}

Next we discuss the key abilities that persistent homology brings to image analysis tasks.
These fall into three general categories: the incorporation of spatial context into a deterministic algorithm, the detection of texture and contrast, and invariance under certain transformations.
\begin{enumerate}
    \item
       {\bf Incorporating spatial context:}
       Many deterministic algorithms, as well as fully-connected neural networks, struggle to incorporate the spatial context inherent in satellite data.
       Integrating this spatial context is precisely what motivated the development of CNNs, but CNNs are costly to train and challenging to make explainable.
       Persistent homology, naturally incorporates spatial context, so patterns that are evident in this spatial context can be incorporated without resorting to CNNs or other spatially-informed neural network architectures.

    \item
       {\bf Detection of texture and contrast:}
       Persistent homology excels at detecting contrast differences -- regions (small or large) that differ from the surrounding average, which gives a representation of the texture present in an image.
       This focus on texture is useful in analyzing satellite weather imagery, as texture is frequently a key distinguishing factor, even more than a cloud being a particular shape or size.
       
    \item   
   {\bf Invariance to homeomorphisms:}
   The notion of not wanting to be constrained by a particular geometry brings us to the final advantage: invariance under a common class of transformations called \emph{homeomorphisms}; see Section~\ref{sec:tda}\ref{subsec:reading_barcodes}.

\end{enumerate}

\subsection{Combining TDA with simple ML algorithms}

\begin{figure}[htp]
    \centering
    \includegraphics[width=.9\textwidth]{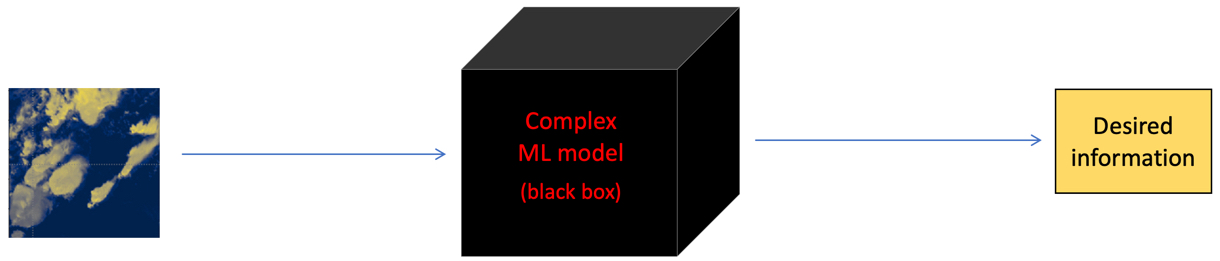}
    
    (a) Pure ML approach: image information extracted using a complex ML model.
    \vspace*{0.5cm}
    
    \includegraphics[width=.9\textwidth]{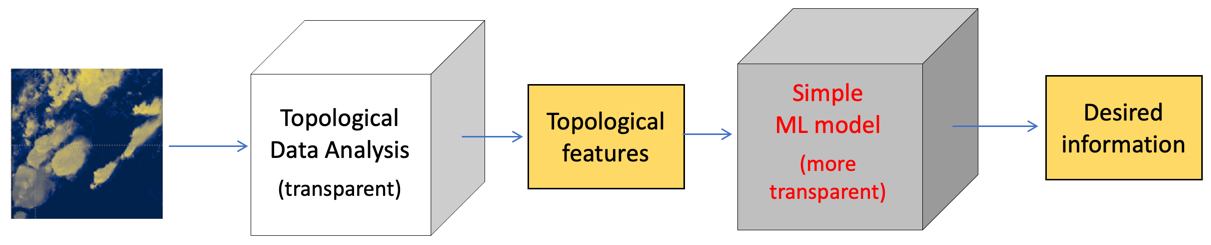}
    
    (b) TDA approach:  image information extracted using TDA followed by simple ML model.
    \vspace*{0.2cm}
    
    \caption{Two different ways to extract desired information from imagery: (a) using a complex ML model, typically a deep neural network; 
    (b) using TDA followed by a simpler machine learning method.
    The latter can lead to more transparent and computationally efficient approaches.
    }
    \label{fig:TDA_ML_graphic}
\end{figure}

For some image analysis tasks TDA methods can be used as a stand-alone tool, but for the majority of tasks, one would first use TDA to extract topological features, then afterwards add a {\it simple} machine learning algorithm, as shown in Figure \ref{fig:TDA_ML_graphic}(b).
For example, the sample application in Section \ref{sec:case_study} uses TDA followed by a support vector machine (SVM).
TDA can thus be viewed as a transparent means to construct new, physically meaningful and interpretable features that may reduce the need for black-box machine learning algorithms.
Using TDA in this way can support the goals of creating ethical, responsible, and trustworthy artificial intelligence approaches for environmental science outlined in \citet{mcgovern2022we}, since transparency is a key requirement for ML approaches to be used in tasks that affect life-and-death decision making \citep{rudin2019stop}, such as severe weather forecasting.

\subsection{Objectives and organization of this article}
\label{subsec:objectives}

As mentioned before, we are not attempting to set a new benchmark for accuracy in classification, nor are we declaring that this method renders existing techniques obsolete.
Instead, we seek to raise awareness of a promising technique with significant potential for ES applications and provide the reader with a high-level understanding of how TDA works, what sorts of questions can be asked using TDA, and how the answers obtained can be interpreted and understood.
The case study in Section~\ref{sec:case_study} provides examples of the sorts of questions TDA can help to address, including reports of negative examples, i.e.\ situations in which persistent homology is {\it not} able to distinguish between classes, which are as informative as positive examples in order to understand the best use of TDA.

The remainder of this article is organized as follows.
Section \ref{sec:guiding_example} discusses in detail the sample application of classifying the mesoscale organization of clouds.
Section \ref{sec:tda} provides an introduction to the key concepts of topological data analysis.
Section \ref{sec:case_study} illustrates the use of these TDA concepts for the sample application from Section \ref{sec:guiding_example}, in combination with a simple support vector machine.
In particular, in Subsection~\ref{sec:case_study}\ref{subsec:interpreting}, we provide a detailed and novel discussion of the characteristic image-level features that our combined TDA-SVM algorithm uses to classify.
This highlights the ability to identify which learned patterns can be exposed and to discuss these in the original feature space, which is one of the greatest strengths of persistent homology and TDA.
Section \ref{sec:advanced} provides an overview of advanced TDA concepts that are beyond the scope of this paper.
Section \ref{sec:conclusion} provides conclusions and suggests future work.

\section{Guiding Application - Classifying the Mesoscale Organization of Clouds from Satellite Data}
\label{sec:guiding_example}

To illustrate the use of TDA we consider the task of identifying patterns of mesoscale (20-20,000km) organization of shallow clouds from satellite imagery, which has recently attracted much attention~\citep{stevens2020sugar,rasp2020combining,denby2020discovering}.
Climate models, due to their low spatial resolution, cannot model clouds at their natural scale~\citep{gentine2018could,rasp2018deep}.
Since clouds play a major role in the radiation budget of the earth \citep{lecuyer2015observed}, the limited representation of clouds in climate models causes significant uncertainty for climate prediction
~\citep{gentine2018could}.
There has been progress in addressing this limitation from the climate modeling side, e.g., using ML to better represent sub-grid processes~\citep{krasnopolsky2005new,rasp2018deep,yuval2020stable,brenowitz2020interpreting}.

\begin{figure}[htp]
    \centering
    \includegraphics[width=.7\textwidth]{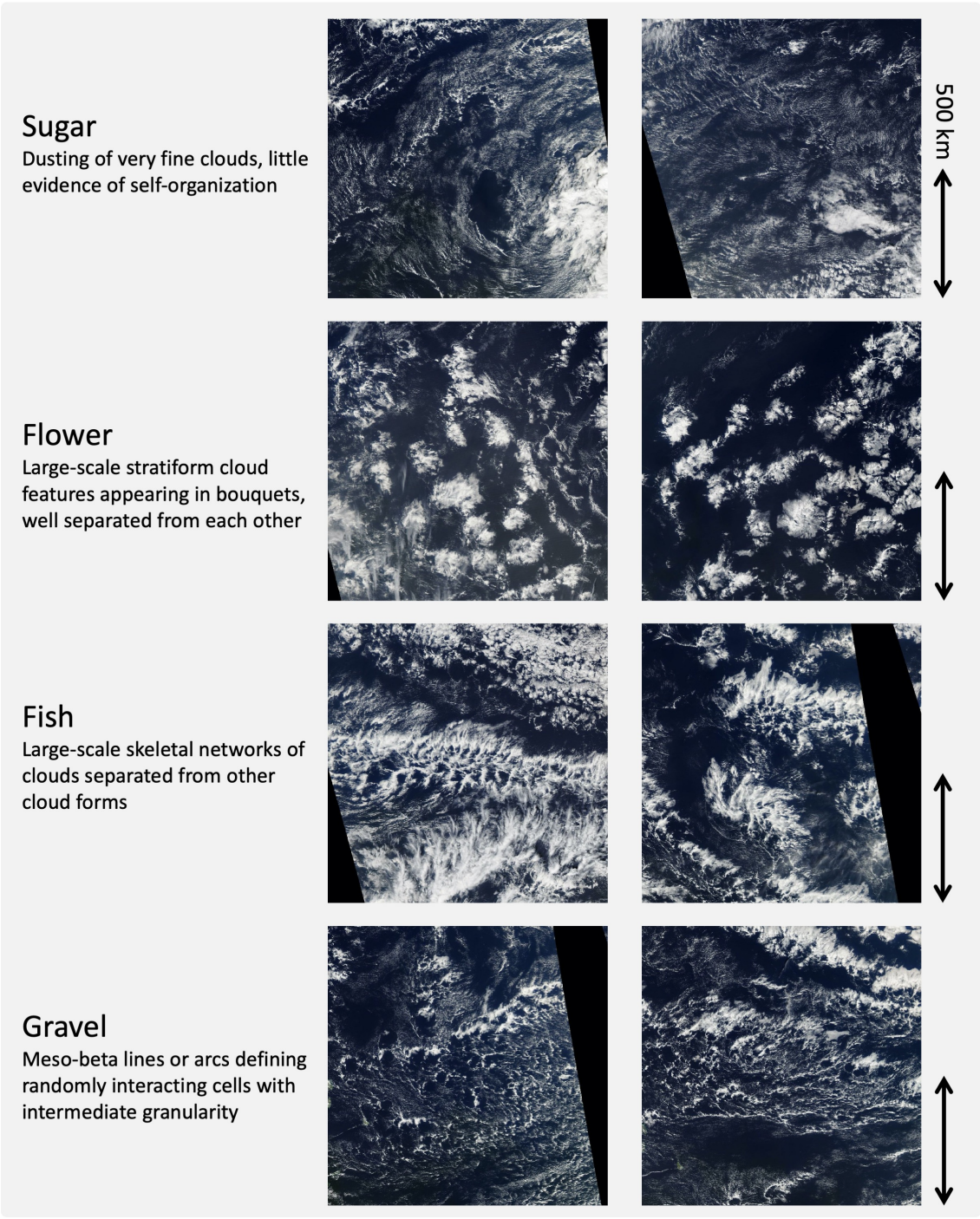}
    \caption{Examples of the four cloud types from the sugar, flowers, fish, and gravel dataset from \citet{rasp2020combining}.
    Note that \citet{rasp2020combining} use the term ``flower'', while we follow \citet{stevens2020sugar} in referring to this type as the plural ``flowers''.
    \emph{Image credit:~Figure 1 in \citet{rasp2020combining}. \copyright~American Meteorological Society. Used with permission.}
    }
    \label{fig:sffg_examples}
\end{figure}

A different approach is to build a better understanding of cloud organization in satellite imagery~\citep{stevens2020sugar,rasp2020combining,denby2020discovering}.
One goal is to track the frequency of occurrence of certain cloud patterns across the globe, reaching back in time as far as satellite imagery allows, to better understand changes to the underlying meteorological conditions.
To this end, in 2020 a group of scientists from an International Space Science Institute (ISSI) International Team identified the primary types of mesoscale cloud patterns seen in Moderate Resolution Imaging Spectroradiometer (MODIS; \cite{gumley2010creating}) True Color satellite imagery, focusing on boreal winter (Dec-Feb) over a trade wind region east of Barbados~\citep{stevens2020sugar}.
Using visual inspection they identified four primary mesoscale cloud patterns, namely sugar, gravel, fish and flowers, shown in Figure~\ref{fig:sffg_examples}.
Subsequent study of these four cloud types using radar imagery~\citep{stevens2020sugar}, and median vertical profiles of temperature, relative humidity and vertical velocity~\citep{rasp2020combining}, indicate that the four cloud types occur in climatologically distinct environments and are thus a good indication of those environments.

While humans are fairly consistent at recognizing these four patterns after some training, it is difficult to describe them objectively so that a machine can be programmed to do the same.
Deep learning offers a potential solution; however, most deep learning approaches require a large number of labeled images to learn from.

\subsection{Approaches for dealing with lack of labeled samples}
\label{subsec:lack_labels}

\citet{rasp2020combining} solve the lack of labeled data for this application by a crowd-sourcing campaign using a two-step process.
First they developed a crowdsourcing environment and recruited experts to label 10,000 images.
Experts used a simple interface to mark rectangular boxes in the imagery and label them with one of the four patterns.
The labeled data set enabled the use of {\it supervised} learning algorithms as a second step, i.e.\ the algorithms were supplied with pairs of input images and output labels and then trained to estimate output labels from given imagery.
Two types of supervised deep learning algorithms were developed, one for object recognition and one for segmentation.
Both algorithms performed well \citep{rasp2020combining}.

In contrast, {\it unsupervised} learning approaches seek to develop models from unlabeled data samples.
Clustering -- which divides unlabeled input samples into groups that are similar in some way -- is a classic unsupervised learning algorithm.
For example, \citet{denby2020discovering} trained an unsupervised deep learning algorithm, in combination with a hierarchical clustering algorithm, for a closely related application, namely grouping image patches from Geostationary Operational Environmental Satellite (GOES; \citet{schmit2017closer}) imagery into clusters of similar cloud patterns.
Their algorithm identified a hierarchy of clustered mesoscale cloud patterns, but since the classes of cloud patterns were generated by a black box algorithm rather than by domain scientists, their meaning is less understood than the four patterns from \citet{rasp2020combining}.
Indeed, a necessary step that comes after the unsupervised learning is to test whether the patterns identified by an algorithm correspond to climatologically distinct environments, and if so which ones.

TDA is an alternative approach to address the lack of labels.
With TDA we seek to match imagery to the original four classes identified by \citet{stevens2020sugar}, yet only require a small number of labeled samples.
We map patches of the MODIS imagery into topological space, then investigate whether there are significant differences in the topological properties that we can leverage to distinguish the patterns.
TDA can thus be viewed as a means of sophisticated feature engineering, giving new, physically meaningful topological features.
Our motivation is that this approach would allow us to identify the well established patterns from~\citet{stevens2020sugar}, but with two key differences: 
(1) this approach does not require a large number of labels (less crowdsourcing required);
(2) this approach is more transparent than the supervised (such as \citet{rasp2020combining}) and unsupervised (such as \citet{denby2020discovering}) deep learning approaches, since topological properties can be understood intuitively.

We note that TDA can also be used in an unsupervised fashion similar to the approach of \citet{denby2020discovering}, only with more transparency and computational efficiency.
On its own, TDA provides an embedding of the image data.
However, rather than the embeddings being a learned property of a neural network whose properties can only be inferred after it is trained, and then only with difficulty, the TDA embedding is deterministically based on topological properties of the image.
For this primer, however, we focus on the supervised task of identifying the previously established patterns of \citet{stevens2020sugar}.

\subsection{Dataset details and preprocessing}
\label{subsec:data_details}

The dataset from~\citet{rasp2020combining} provides approximately 50,000 individual cloud type annotations on around 10,000 base images.
In~\citet{rasp2020combining} comparison of Intersection over Union (IoU) scores, also known as Jaccard index \citep{jaccard1901etude,fletcher2018comparing},
between annotators analyzing the same image indicated that these crowd-sourced annotations were of generally high quality.
See Figure~\ref{fig:sffg_examples} for examples of these cloud types; in general, sugar type clouds are small, relatively uniformly distributed clouds; gravel type clouds are somewhat larger than sugar clouds, and tend to show more organization; flowers type clouds are yet larger clouds that clump together with areas of clear sky between; and fish type clouds form distinctive mesoscale skeletal patterns.
Each image in the dataset is a $14^\circ \times 21^\circ$ (lat-lon) visible-color MODIS image from the Terra or Aqua satellite.
On these images, annotators could draw rectangular annotations encompassing a single cloud type, and could apply as many annotations to each image as they desired, so long as each annotation encompassed at least 10\% of the image.

As we will discuss later, persistent homology takes as its input a space with an intensity value at each point, which in our case corresponds to a grayscale image.
The MODIS images in the dataset from~\citet{rasp2020combining} were NASA Worldview True Color images in RGB \citep{gumley2010creating}, which we converted to grayscale using the python package \verb|pillow|, which uses the ITU-R BT.601-7 luma transform~\citep{itur2011} for computing intensity from RGB input:
\[
    I = 0.299 R + 0.587 G + 0.114 B.
\]
This is a transform originally developed for television broadcasting and approximates the overall perceived brightness for each pixel, which is appropriate here as the NASA Worldview True Color images are a close approximation of what a human observer in orbit would see.
We note that this is a difference between our work and that of \citet{rasp2020combining}, as they used the RGB images throughout.

\section{Introduction to Topological Data Analysis} 
\label{sec:tda}

In this section we provide a brief introduction to relevant mathematical topics.
For more details, we refer readers to~\citet{carlsson2009topology,ghrist2008barcodes} and \citet{edelsbrunner2010computational}.

\subsection{Topology}
\label{subsec: topology}

In the broadest sense, topology is the study of the fundamental shapes of abstract mathematical objects.
When we speak of the ``topology'' of an object, we speak of properties that do not change under a smooth reshaping of the object, as if it is made of a soft rubber.
Some example properties include: how many connected components the object contains, how many holes or voids it contains, and in what ways the object loops back on itself.
In this paper, we focus on the first two properties: connectivity and holes.

\subsection{Homology}
\label{subsec:homology}

\begin{figure}[htp]
    \centering
    \includegraphics[width=.7\textwidth]{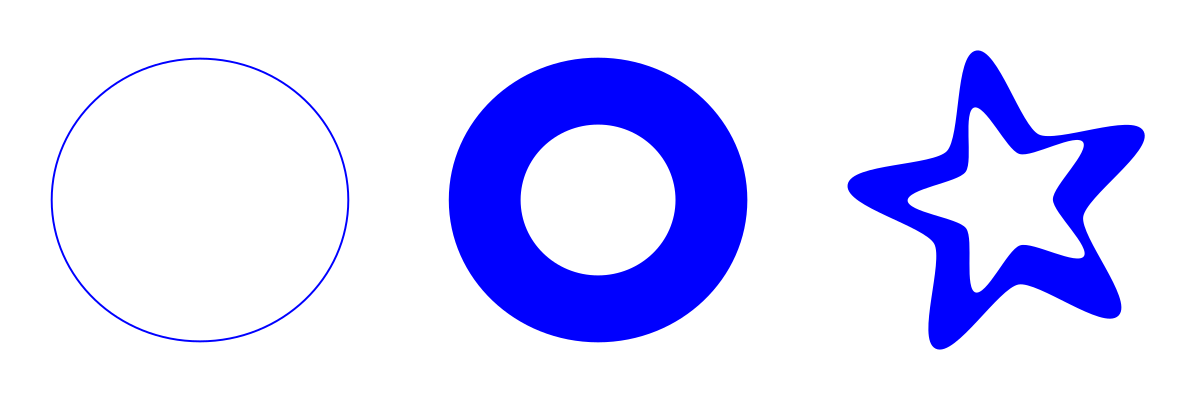}
    \caption{Three shapes that each have the same homology -- a single connected component, a single 1-dimensional hole, and no higher-dimensional holes.}
    \label{fig:hom_ex}
\end{figure}

Homology is one of the tools from topology that focuses on connectivity and holes.
The $d$-dimensional homology $H_d$ (for $d \in \Z_{\ge 0}$) counts the number of $d$-dimensional holes (or voids) in that object.
For $d=0$, the 0-dimensional homology $H_0$ captures the number of connected components present in an object.
For $d \ge 1$, the homology $H_d$ captures holes -- a 1-dimensional hole is one that can be traced around with a 1-dimensional loop (like a loop of string), while a 2-dimensional hole is a void.
As shown in Figure \ref{fig:hom_ex}, these holes and the surrounding surface need not be circular -- they may be deformed along the lines of Figure \ref{fig:homeomorphism_ex}.
Because homology is only interested in counting the presence of these features, it is invariant under any transformation of the space that does not create or destroy any holes or components.
In our application of grayscale images, no holes of dimension 2 or larger can exist, as that would require a dataset that is at least 3-dimensional.

\subsection{Persistent homology}
\label{subsec:pers_hom}

While homology focuses on global features of the space, there is an extension, known as \emph{sublevelset/superlevelset persistent homology}, that captures more small-scale geometry~\citep{edelsbrunner2010computational}.
Superlevelset persistent homology is the primary tool from TDA we use in this paper, as it gives the best descriptor of image texture.

\begin{figure}[htp]
    \centering
    \includegraphics[width=\textwidth]{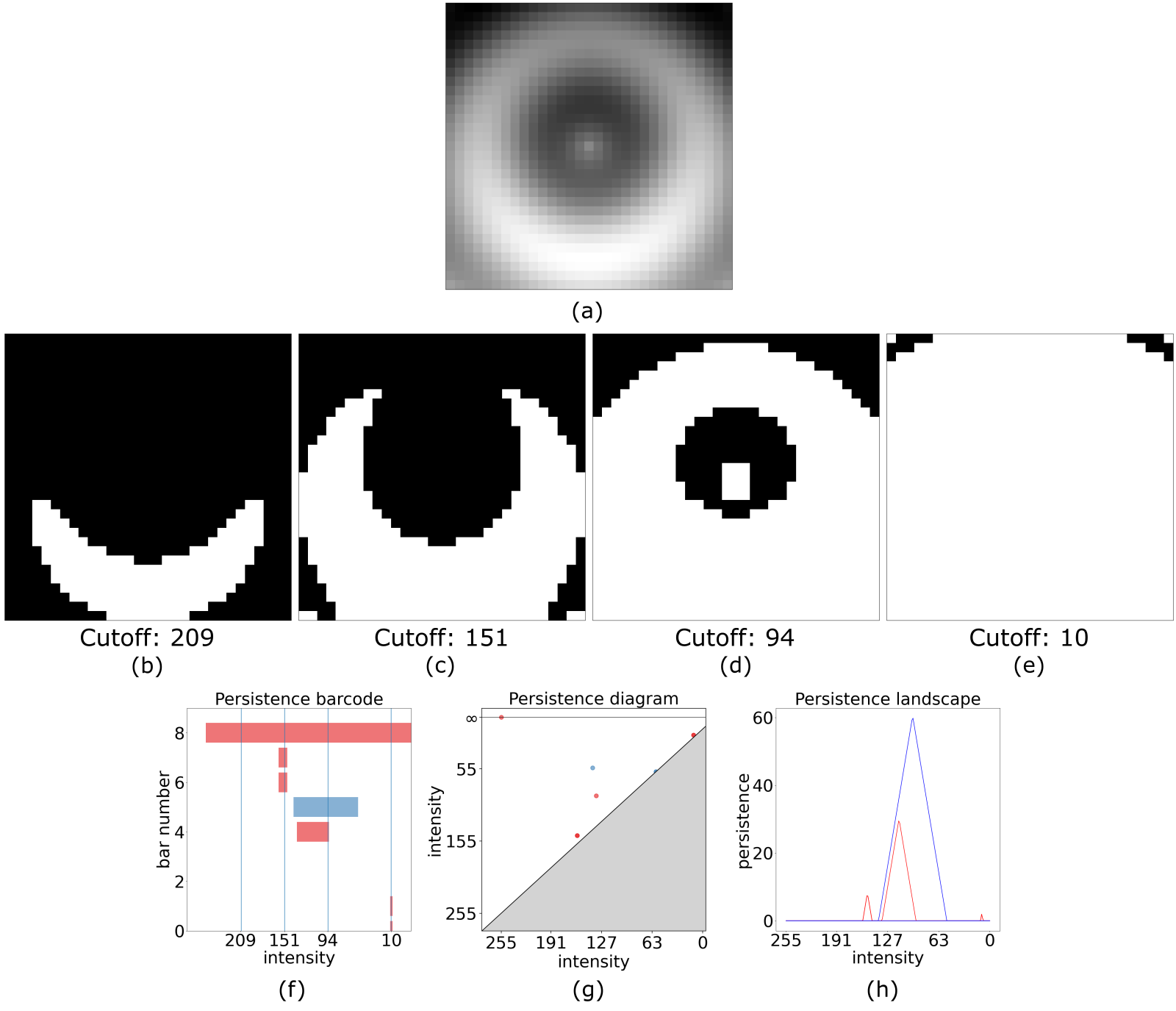}
    \caption{
        A grayscale image (a) ranging between intensity 0 for black and 255 for white.
        Four superlevelsets from (a) at cutoff values 209, 151, 94, and 10 areshown in (b) through (e), respectively.
        The pixels included in the superlevelsets are colored white.
        The persistence barcode for (a) is in (f), with vertical lines indicating the intensities corresponding to the four superlevelsets in (b) through (e).
        The equivalent persistence diagram is in (g), and (h) is the resulting persistence landscape.
        In the barcode, diagram, and landscape, red elements (bars, points, and lines, respectively) indicate connected components ($H_0$ features), while blue elements indicate 1-dimensional holes ($H_1$ features).
    }

    \label{fig:sub_hom_ex}
\end{figure}

For an example of superlevelset persistent homology being computed on a simple surface, see Figure~\ref{fig:sub_hom_ex}.
The input to superlevelset persistent homology is a $d$-dimensional space plus an intensity value at every point.
In our example, this is a grayscale image with two spatial dimensions ($d = 2$) with the pixel values as intensities.
This input is then converted into \emph{superlevelsets}: each superlevelset is a binary mask of the original space, in which only points that have an intensity value {\it greater} than a particular cutoff value have been included.
As this cutoff value sweeps down from the maximum intensity, the homology of each superlevelset is computed and the cutoff values at which homological features (connected components, holes) appear and disappear are tracked.

For our example, this means that as the cutoff value decreases, more and more pixels with gradually decreasing intensities are included in the superlevelsets, and we track the connected components and holes that appear and disappear.
Because we are using superlevelsets, in which we start by including the highest intensities, we can view connected components appearing at high intensity value as being analogous to cloud tops, which are typically brighter, and holes as darker regions within these bright clouds.

This added interpretability motivates our focus here on {\it superlevelset persistent homology}, which are a simple variation (reflection) of the more commonly used {\it sublevelset persistent homology}.
Sublevelset persistent homology is computed exactly the same way, but instead of each set including all the pixels with intensities above the cutoff value, pixels with intensities below the cutoff value are included, and the cutoff value is viewed as sweeping from low intensities up to high intensities.
Throughout this paper, we will frequently omit the prefix ``superlevelset'' in ``superlevelset persistent homology'' and simply use ``persistent homology'' to refer to this technique -- this should not be confused with the persistent homology technique which takes as its input a cloud of data points \citep{carlsson2009topology}.

In practice, it is not necessary to compute the homology for infinitely many superlevelsets -- there are algorithms which discretize the data and then use linear algebra to implement this computation efficiently.
These implementations are fast for low-dimensional data (e.g., the $2$-dimensional grayscale images used in our guiding example) but become more resource-intensive as the dimension of the input space grows.

\subsection{Persistence barcodes and diagrams}
\label{subsec:barcodes-diagrams}

There are two main ways to display the output of persistent homology: persistence barcodes (Figure~\ref{fig:sub_hom_ex}f) and persistence diagrams (Figure~\ref{fig:sub_hom_ex}g).

In a barcode, each homological feature that appears is represented by a horizontal bar, which stretches from the cutoff value at which the corresponding feature first appears (is \emph{born}) to the value at which it disappears (\emph{dies}).
Because we are using superlevelset persistent homology, our cutoff values are decreasing; thus, the intensity values on the $x$-axis are decreasing from left to right.
The \emph{persistence} of each feature is the length of its bar.
To distinguish between different homological classes, we color the bars depending on what dimension the homological feature is.
We use red bars for 0-dimensional features (connected components), and blue bars for 1-dimensional features (holes).
The $y$-axis of a persistence barcode counts the number of bars, typically ordered by birth value.

A persistence diagram contains the same information as a persistence barcode, but represents each feature as a point rather than as a bar.
In persistence diagrams, both the $x$-axis and $y$-axis represent intensity.
The $x$-coordinate of this point is given by the birth cutoff value, while the $y$-coordinate is the death cutoff value.
Because features always die at a higher cutoff value than they are born, all points lie above the diagonal line $y = x$.
The persistence of a feature is represented by how far a persistence diagram point lies above the diagonal.
We present persistence diagrams here to familiarize the reader with their use in, e.g., \citet{kim2019deciphering, tymochko2020using, sena2021topological}, but for our case study we focus on barcodes and landscapes.

In persistent homology, there are features that have infinite persistence -- features which are born at a particular intensity, but never die.
The most common example of this is that the first connected component to appear will eventually become the only remaining connected component, as all other components eventually merge into it at high enough cutoff values.
These infinite-persistence points are represented as infinite bars (rays) in persistence barcodes, stretching out of the frame to the right, and as infinite points appearing on a special ``$+\infty$'' line in persistence diagrams.

\subsection{How to read and interpret persistence barcodes}
\label{subsec:reading_barcodes}

To demonstrate how to read a persistence barcode we return to Figure~\ref{fig:sub_hom_ex}.
The four vertical lines in the barcode (f) correspond to the four superlevelsets in the middle row, where white pixels show regions included in the superlevelset.
In (b), we see the first connected component appear, corresponding to the top, infinite-length red bar in the barcode.
In (c), two small components in the lower corners appear, corresponding to the two short red bars in the barcode crossed by the second vertical line.
The short length of these bars indicates that these components are short-lived, and soon merge into the larger component.
In (d), we see both the central 1-dimensional hole, which corresponds to the blue bar, and the connected component within that hole which corresponds to the red bar that is about to end near the third vertical line in the barcode.
Finally, in (e), we see almost the entire image is in the superlevelset, as the cutoff value is very small.
However, the upper two corners have just been included as two new components, which are even more short-lived than the lower corner components, as indicated by their extremely short red bars.

\begin{figure}[htp]
    \centering
    \includegraphics[width=.8\textwidth]{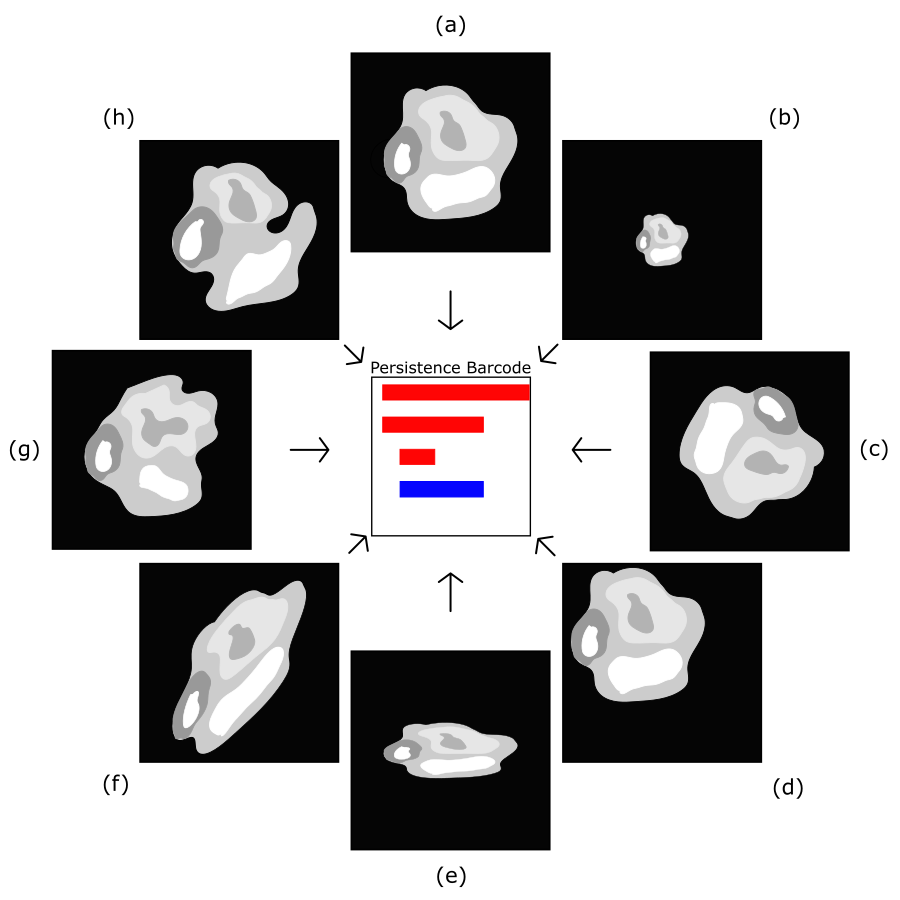}
    \caption{
        Examples of deformations that all result in the same persistence barcode.
        The original image is shown in (a), and the transformations are as follows: scaling in (b), rotation in (c), translation in (d), uneven scaling in (e), shearing in (f), and general homeomorphisms in (g) and (h).
    }
    \label{fig:homeomorphism_ex}
\end{figure}

The first thing to notice about this barcode is that there are relatively few red bars, apart from the infinite-length bar, and those bars are quite short.
This indicates that few connected components appear and disappear as we scale through intensity values, and thus the base image is quite smooth.
There is one red bar of reasonable length, so we would expect there to be one somewhat significant ``bump'', a bright region surrounded by darker regions, which is precisely what we see in the middle of Figure~\ref{fig:sub_hom_ex}a.
We also notice that there is one relatively long blue bar, which tells us that there is a hole (dark region surrounded by brighter regions) which persists for a relatively wide range of intensities.

Persistent homology is invariant under \emph{homeomorphisms} of the input space, which are continuous deformations with continuous inverses; see Figure~\ref{fig:homeomorphism_ex}.
Examples include all the rigid motions of the plane (rotation and translation), affine transformations (scaling, skewing, etc), as well as more radical reshapings, so long as no ``ripping'' occurs.
Superlevelset persistent homology is invariant over all such transformations.
So, a cloud that has been reshaped, expanded, and moved but which retains the same overall texture as in its original incarnation would have the same superlevelset persistence barcodes.
See Section~\ref{sec:advanced} for some brief comments on versions of persistent homology that can distinguish between such different deformations of an image.

\subsection{Persistence landscapes}
\label{subsec:pers_lscapes}

Persistence barcodes and diagrams have a drawback: they are not always convenient inputs for use in machine learning tasks, as described by~\citet{bubenik2015statistical,adams2017persistence} and \citet{mileyko2011probability}, since they do not naturally live in a vector space.
To deal with this, we use \emph{persistence landscapes} to summarize and vectorize the persistence diagram~\citep{bubenik2015statistical}.
A persistence landscape is a collection of piecewise-linear functions that capture the essence of the persistence diagram.

\begin{figure}[htp]
    \centering
    \includegraphics[width=\textwidth]{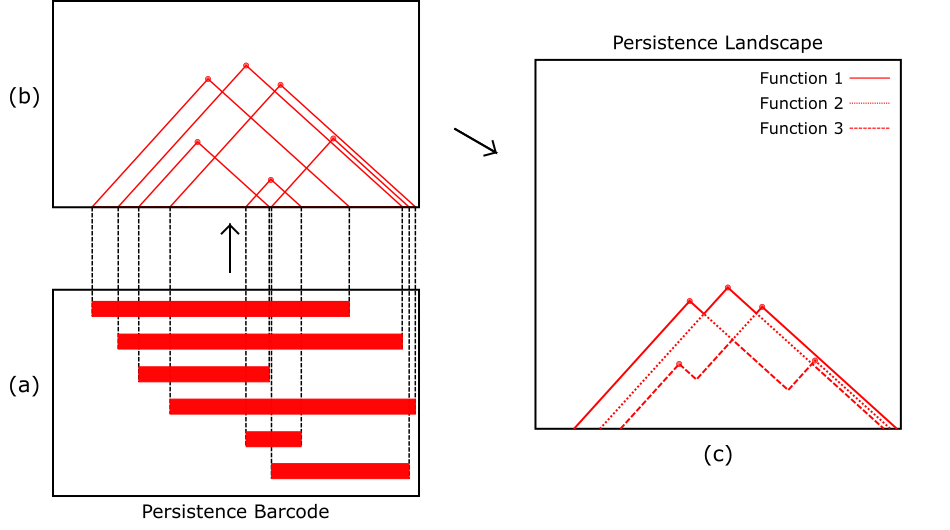}
    \caption{
        The process of computing a persistence landscape from a barcode.
        Beginning with the barcode in (a) (which already has the infinite bar removed), we raise a ``mountain'' above each bar to obtain the ``mountain range'' in (b).
        Our first persistence landscape function is the piecewise-linear function that follows the highest edges of the mountain range in (b), shown as the solid line in (c).
        The next function in (c) is obtained by deleting in (b) the lines corresponding to this first function, then finding the piecewise-linear function that follows the highest edges of this modified mountain range in (b) -- this is the dotted line in (c).
        Further landscape functions are computed similarly, by deleting previous functions in (b) and tracing along the highest remaining edges.
        In (c), only the first three landscape functions are shown.
    }
    \label{fig:landscape_ex}
\end{figure}

An example of a persistence landscape computed from a small persistence barcode is shown in Figure~\ref{fig:landscape_ex}.
We separate out a particular homological dimension (e.g., $H_0$ or $H_1$) and remove any infinite bars, then create a new figure containing a collection of isosceles right triangles with hypotenuses along the $x$-axis, one for each bar in the barcode.
These triangles are scaled so that the triangle corresponding to a bar is the same width as that bar.
We view this collection of triangles as a ``mountain range'', and begin to decompose it into landscape functions.
The first landscape function is the piecewise-linear function that follows the uppermost edge of the union of these triangles, i.e., it is the top silhouette of the mountain range.
To compute the next landscape function, we delete the first landscape function from the mountain range, then find the piecewise-linear function that follows the uppermost edge of this new figure at every point, and so forth for the further landscape functions.
This collection of piecewise-linear functions is the persistence landscape.
The $x$-axis still represents intensity, while the height of each peak is proportional to the length of the bar it came from, and is thus a measure of persistence.

This representation is stable --- small changes to the input will only result in small changes in the persistence landscape~\citep{bubenik2015statistical}.
While the entire persistence landscape determines a persistence barcode exactly, in our work we retain only the top several persistence landscape functions, which means that we obtain a descriptive summary capturing the information of high-persistence points, but ignore some information about low-persistence points.

\subsection{How to read and interpret persistence landscapes} 
\label{subsec:reading_lscapes}

\begin{figure}[htp]
    \centering
    \includegraphics[width=\textwidth]{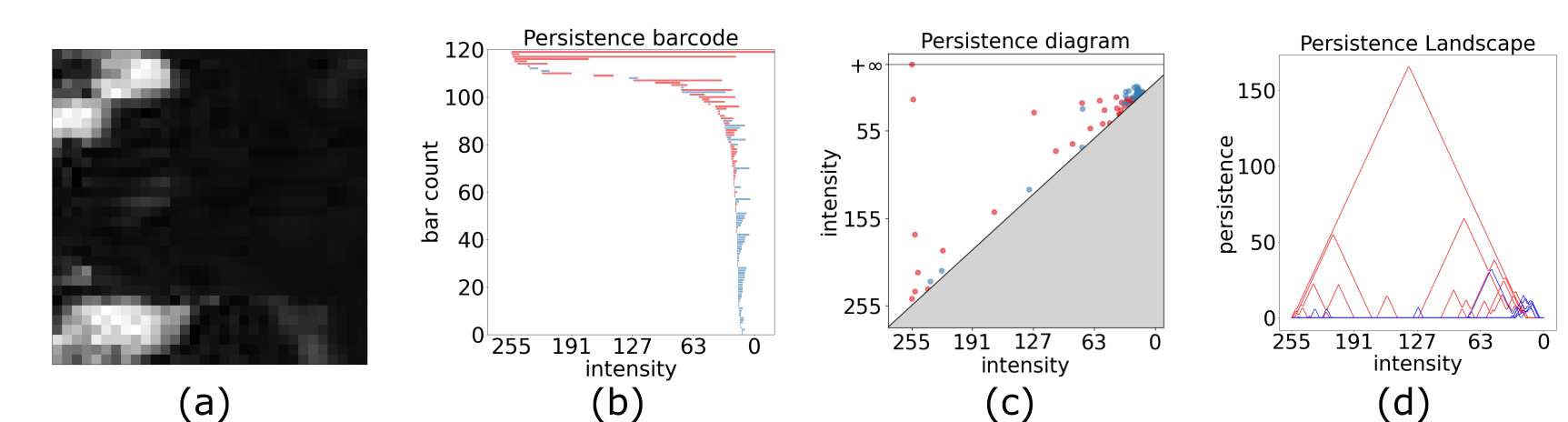}
    \caption{(a) A $32 \times 32$ sample image from the grayscale MODIS imagery used in the sugar, flowers, fish, and gravel dataset, along with (b) its persistence barcode, (c) persistence diagram, and (d) persistence landscape with the first 5 piecewise-linear functions for each of the $H_0$ and $H_1$ classes.}
    \label{fig:cloud_lscape_ex}
\end{figure}

We now look at a realistic example in Figure~\ref{fig:cloud_lscape_ex}.
Reviewing the barcode in \ref{fig:cloud_lscape_ex}b, we first notice two red bars with high persistence -- the infinite bar, as well as another that stretches nearly all the way across the barcode.
This indicates that there are two high intensity regions that are separated by a dark region.
Over middling intensities, there are few bars, indicating that outside the two bright regions we already identified, there is little going on.
Finally, when we get to lower intensities (darker regions), there are many short bars, representing subtle variations in the dark regions.
This information, however, is somewhat hard to read in a barcode with as many bars as this.
Thus, we turn to landscapes as a way to summarize this information in a more readable format.

In the landscape in Figure~\ref{fig:cloud_lscape_ex}d, the single tall red mountain indicates that the original image (Figure~\ref{fig:cloud_lscape_ex}a) contains two connected components that persist over a large range of intensity levels (as the infinite-persistence component is implicit in the landscape).
The only blue mountains in the landscape are much smaller than the red mountains and are mainly to the right of them.
This indicates that while the original image contains numerous holes within the connected components, they only appear near the bottom end of the intensity range, i.e.\ the holes do not appear until we have started including relatively dark regions.
As an example, consider the single extremely dark pixel in the upper left-hand corner adjacent to the bright clouds, and surrounded by a moderately dark region.
This hole contributes a moderately tall blue peak far to the right in the landscape, as it does not appear until the relatively dark region surrounding it is included into the bright adjacent component, but will not fill in until the nearly black pixel in the middle of the hole is included.

In general, high-persistence features (long bars, tall mountains) give information about large-scale features -- the presence of two bright clouds in Figure~\ref{fig:cloud_lscape_ex}a, for example.
On the other hand, low-persistence features (short bars, small mountains) give information about texture.
In~\ref{fig:cloud_lscape_ex}, the short bars and small mountains appearing at lower intensity values indicate that the background darkness in the image is relatively noisy, rather than being uniformly black or smoothly graded.
The few small blue mountains in Figure~\ref{fig:cloud_lscape_ex}d indicate that the bright clouds also contain some textural elements - regions of slightly darker cloud within brighter regions.

\section{Environmental Science Satellite Case Study} 
\label{sec:case_study}

Now that we have established the basic theory of TDA, we return to its application to classifying mesoscale clouds.

\subsection{Adapting persistent homology to this dataset} 
\label{subsec:adapting_persistence}

\begin{figure}[htp]
    \centering
    \includegraphics[width=.9\textwidth]{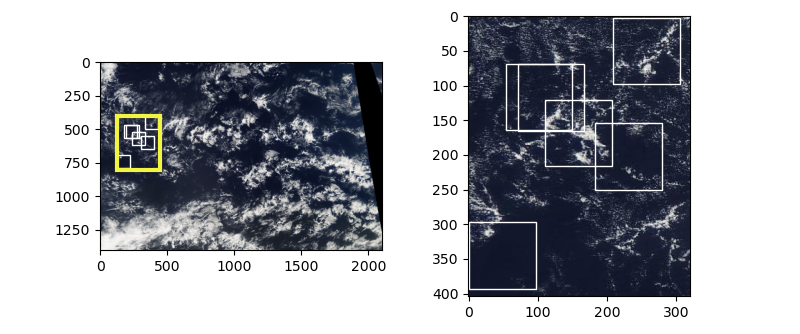}
    \caption{On the left, a full $14^\circ \times 21^\circ$ example image is shown, with an example sugar annotation shown in yellow.
        This annotated region is shown in detail on the right.
        In both images, the six $96 \times 96$ pixel subsamples are denoted by white squares.}
    \label{fig:annotation_subsamples}
\end{figure}

\begin{figure}[htp]
    \centering
    \includegraphics[width=\textwidth]{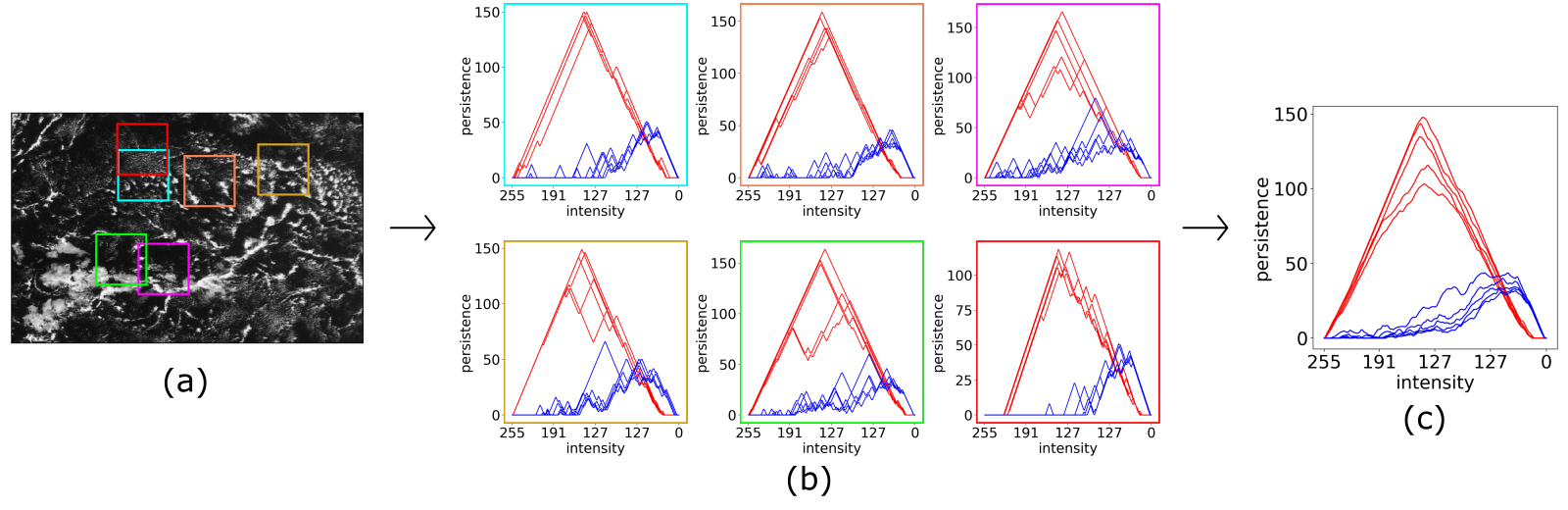}
    \caption{An illustration of the subsampling process.
In (a), we see the full annotation with the color-coded subsamples.
In (b), the 6 raw landscapes with bounding boxes colored to match the subsamples in (a) are displayed, and in (c), the resulting averaged landscape is displayed.}
    \label{fig:subsampling_cartoon}
\end{figure}

While the overall images in the dataset of \citet{rasp2020combining} are of consistent size ($1400 \times 2100$ pixels), the annotations surrounding distinct cloud regimes are not.
An annotation covering more area has inherently more complexity, which would yield a barcode with more bars than a smaller annotation.
To account for this, we implemented a subsampling routine.
For each annotation, six $96 \times 96$ pixel subsamples were randomly selected and their persistence landscapes computed.
The $H_0$ and $H_1$ landscapes each consisted of five functions, which were each sampled at 200 evenly-spaced points.
These samples were then concatenated, yielding a real-valued vector of dimension 2,000 (i.e., a point in $\R^{2000}$) for each subsample.
These six points in $\R^{2000}$ were averaged to obtain a single embedding for the annotation.
This point can also be displayed and interpreted as a landscape, in which (for instance) the top $H_0$ function is obtained by taking the average the corresponding top $H_0$ functions in the 6 subsamples.
An example of the full image, annotated region, and subsamples is shown in Figure~\ref{fig:annotation_subsamples}, and an illustration of the subsampling process is shown in Figure~\ref{fig:subsampling_cartoon}.

\subsection{Dimensionality reduction and adding a simple machine learning model to build a classifier}
\label{subsec:ml}

Once we obtained vectorized representations of each annotation, we sought to visualize the dataset.
As $\R^{2000}$ is not visualizable, we applied a dimensionality reduction algorithm to yield a representation that we can plot.
We used principal component analysis (PCA), as it is a widely-used and relatively simple technique, which in our case produced quite good results.
We found that patterns in the data were visible upon projecting down onto the first three principal components, which captured over 90\% of the variation in the high-dimensional data.
We also note that the principal component vectors from repeated random samplings were extremely consistent, indicating that our projections were quite stable.

Once the data were projected down to three dimensions, we could visualize the data as a point cloud, with points colored according to which cloud pattern they represent.
As a note, the PCA algorithm was entirely unsupervised with regards to these cloud pattern labels -- it used only the vectorized representation of the persistence diagram.

We analyzed each of the six pairs of classes separately by training a support vector machine (SVM, \citet{boser1992training}) to find the plane that best separates the two classes of projected data in three dimensions.
We considered running the SVM on the high-dimensional data, but initial testing indicated that this tended to overfit the data, and actually resulted in reduced classification performance.
The SVM was trained on a random sample of 350 annotations of each class, then performance metrics were computed for a test set of 200 random annotations of each class.
For visualizations of the test data and SVM separating plane, as well as performance metrics for each of the six pairwise comparisons, see Figure~\ref{fig:3D_plots}.

\subsection{Results - what initial patterns emerged from applying persistent homology?} 
\label{subsec:initial_patterns}

\begin{figure}[htp]
    \centering

    \begin{subfigure}{0.48\textwidth}
        \centering
        \includegraphics[width=\textwidth]{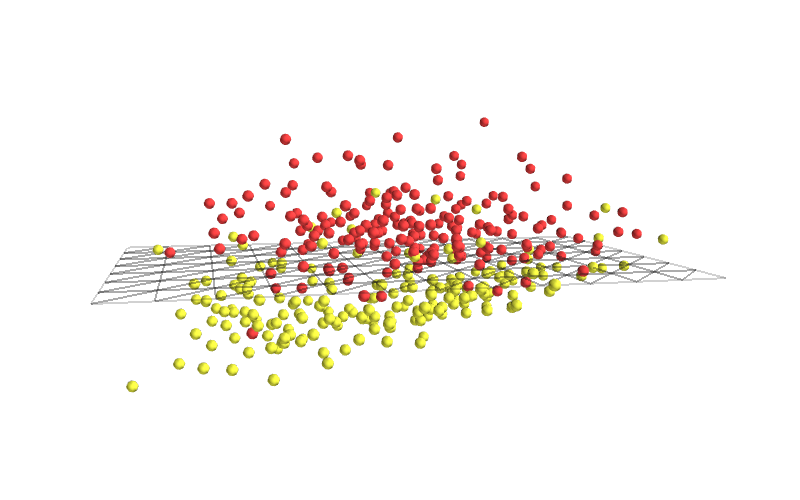}
        \caption{Sugar vs.\ flowers.
        Test data performance: 89.25\%.}
        \label{subfig:sugar_flowers_3D}
    \end{subfigure}
    \hfill
    \begin{subfigure}{.48\textwidth}
        \centering
        \includegraphics[width=\textwidth]{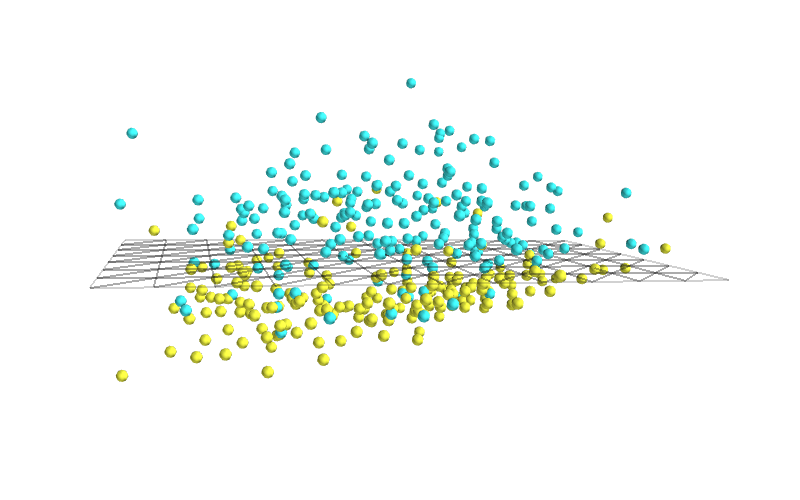}
        \caption{Fish vs.\ sugar.
        Test data performance: 86\%.}
        \label{subfig:fish_sugar_3D}
    \end{subfigure}

    \medskip 

    \begin{subfigure}{0.48\textwidth}
        \centering
        \includegraphics[width=\textwidth]{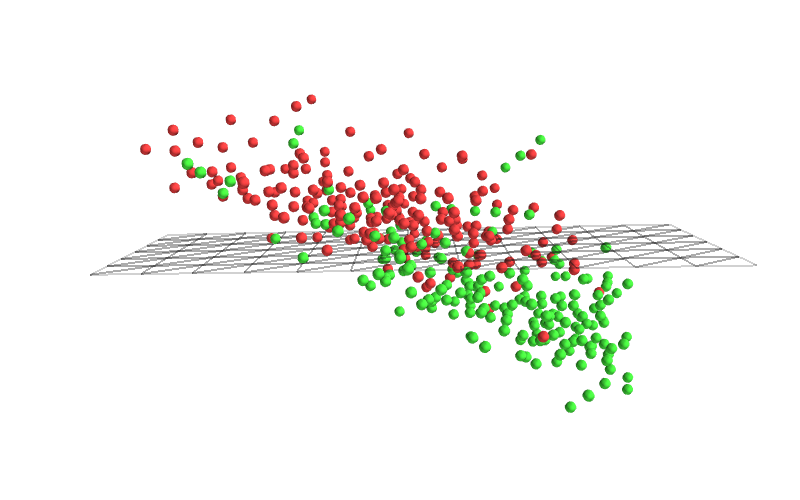}
        \caption{Flowers vs.\ gravel.
        Test data performance: 81\%.}
        \label{subfig:flowers_gravel_3D}
    \end{subfigure}
    \hfill
    \begin{subfigure}{.48\textwidth}
        \centering
        \includegraphics[width=\textwidth]{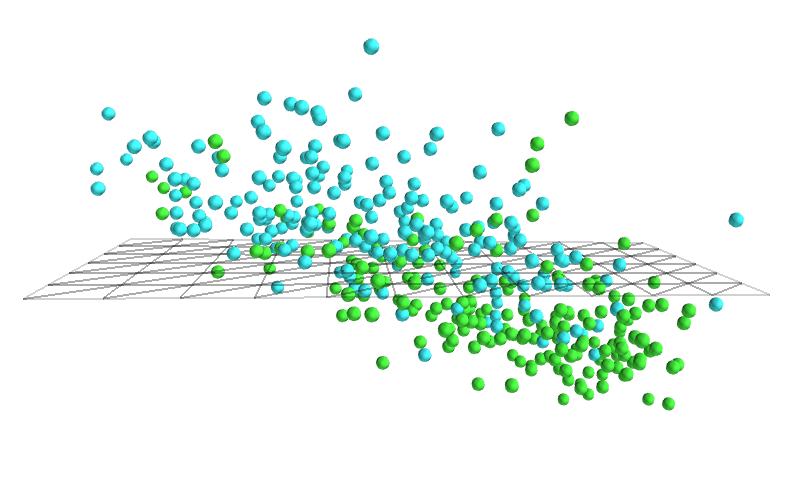}
        \caption{Fish vs.\ gravel.
        Test data performance: 79.5\%.}
        \label{subfig:fish_gravel_3D}
    \end{subfigure}

    \medskip 
    \begin{subfigure}{0.48\textwidth}
        \centering
        \includegraphics[width=\textwidth]{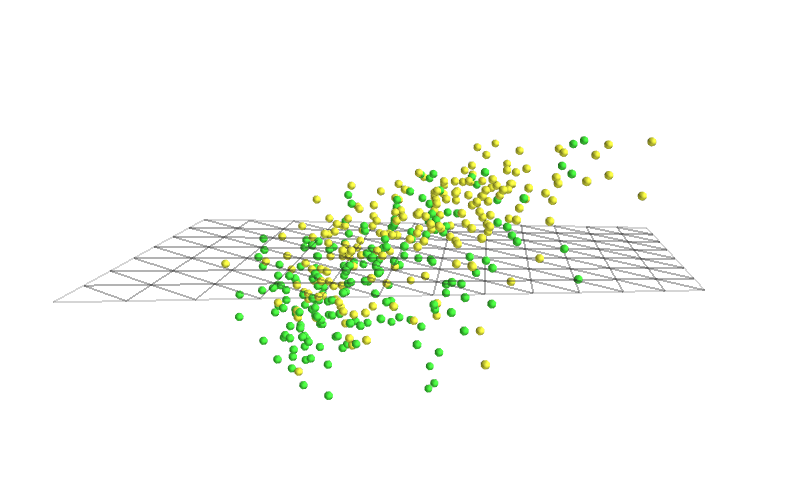}
        \caption{Sugar vs.\ gravel.
        Test data performance: 71.25\%.}
        \label{subfig:sugar_gravel_3D}
    \end{subfigure}
    \hfill
    \begin{subfigure}{.48\textwidth}
        \centering
        \includegraphics[width=\textwidth]{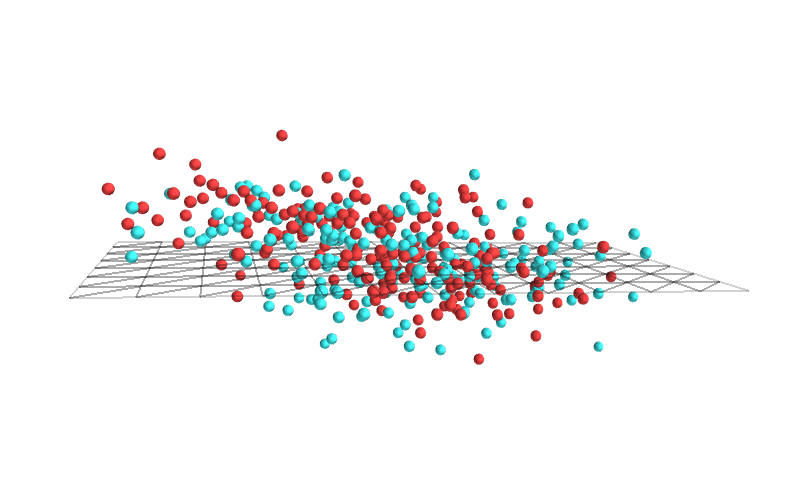}
        \caption{Fish vs.\ flowers.
        Test data performance: 57\%.}
        \label{subfig:fish_flowers_3D}
    \end{subfigure}

    \caption{Plots of the embedded landscape test points for each pairwise combination.
    Flower points are red, sugar points are yellow, gravel points are green, and fish points are blue, with color representing the class assigned in the crowdsourced data set in \citet{rasp2020combining}.
    The SVM separating plane for each pair is shown in wireframe, and the percentage of correctly classified points is reported.}
    \label{fig:3D_plots}
\end{figure}

Overall, the projected points separated well, with one notable exception.
We began our investigation by comparing the sugar vs.\ flowers patterns, as these are visually the most dissimilar (Figure~\ref{fig:sffg_examples}).
As expected, these classes had the most distinct separation out of the six possible pairs, as seen in Figure~\ref{subfig:sugar_flowers_3D}.
While the separation was not perfect, most of the error comes in the form of some intermingling near the separating hyperplane.
The performance of the algorithm in separating these classes was striking, given that only a small, random subset of each annotation was included, and that the PCA projection algorithm was entirely blind to the data labels.
This example is a clear indicator of the potential that persistent homology has to usefully extract textural and shape differences in satellite imagery.

While not quite as exceptional, the flowers and gravel patterns also separate well, as seen in Figure~\ref{subfig:flowers_gravel_3D}.
There is again a degree of intermingling near the separating plane, and in this case that intermingling extends a bit farther to either side.
This is what we would expect, based on the visual presentation of the cloud regimes -- the gravel class falls somewhere between sugar and flowers in terms of cloud size and organization.

We begin to see the algorithm struggle a bit more when we attempt to separate the sugar regime from the gravel regime.
As we can see in Figure~\ref{subfig:sugar_gravel_3D}, the intermingling of data points stretches throughout the point cloud, although there is still a difference in densities between the classes on either side of the separating hyperplane.

However, there are two classes that are effectively indistinguishable by this algorithm: the flowers and fish patterns.
The plot of these points can be seen in Figure~\ref{subfig:fish_flowers_3D}, and it is apparent that there is no effective linear separator between these classes.
While there is a ``separating'' hyperplane plotted, it is much less relevant in this case than in the others; the data points are remarkably evenly-mixed.
A potential explanation for why the algorithm struggles so much with this task is that the distinguishing features of fish vs.\ flowers are simply too large-scale for the subsampling technique to pick up on.
The fish pattern is characterized by its mesoscale skeletal structure, particularly in its difference from the flowers regime, which is more randomly distributed.
This mesoscale organization is simply not visible to the subsamples, as the $96 \times 96$ patches are too small to detect that skeletal structure.
We also note that in \citet{rasp2020combining}, the fish pattern was the most controversial amongst the expert labelers, so it is perhaps not surprising that our algorithm also struggles.

When we look at fish vs.\ sugar and fish vs.\ gravel in Figures~\ref{subfig:fish_sugar_3D} and~\ref{subfig:fish_gravel_3D} respectively, we can see how similar these plots appear to those in Figures~\ref{subfig:sugar_flowers_3D} and~\ref{subfig:flowers_gravel_3D}, in which flowers was compared with sugar and gravel.
This similarity is made even more remarkable by the fact that the sugar samples in these plots were drawn separately rather than being reused for the pairwise comparisons (and similarly for the gravel samples).
While the algorithm is not doing well at distinguishing between fish and flowers, we can at least see that its behavior is consistent: fish and flowers are projected similarly into the 3D embedding space, so they compare similarly with the other classes.

Overall, this case study suggests that it is possible to use persistent homology to quantify and understand the shape and texture of satellite cloud data.
While there are cases where the algorithm struggles, these are understandable in terms of the visual task being requested, and are internally consistent from sample to sample.
Moreover, in the cases where the algorithm does well, it does so consistently across repeated samples, and suggests that when these tools are appropriately applied, excellent results can be obtained from very limited sample sizes.

\subsection{A novel interpretation method -  deriving interpretations in terms of weather and homology} 
\label{subsec:interpreting}

As an example of how this separation can be interpreted, we examine the case of sugar vs.\ flowers.
Recall that in Figure~\ref{subfig:sugar_flowers_3D}, we saw that this pair of classes had the strongest separation in the dataset.

\begin{figure}[htp]
    \centering

    \begin{subfigure}{0.3\textwidth}
        \centering
        \includegraphics[width=.85\linewidth]{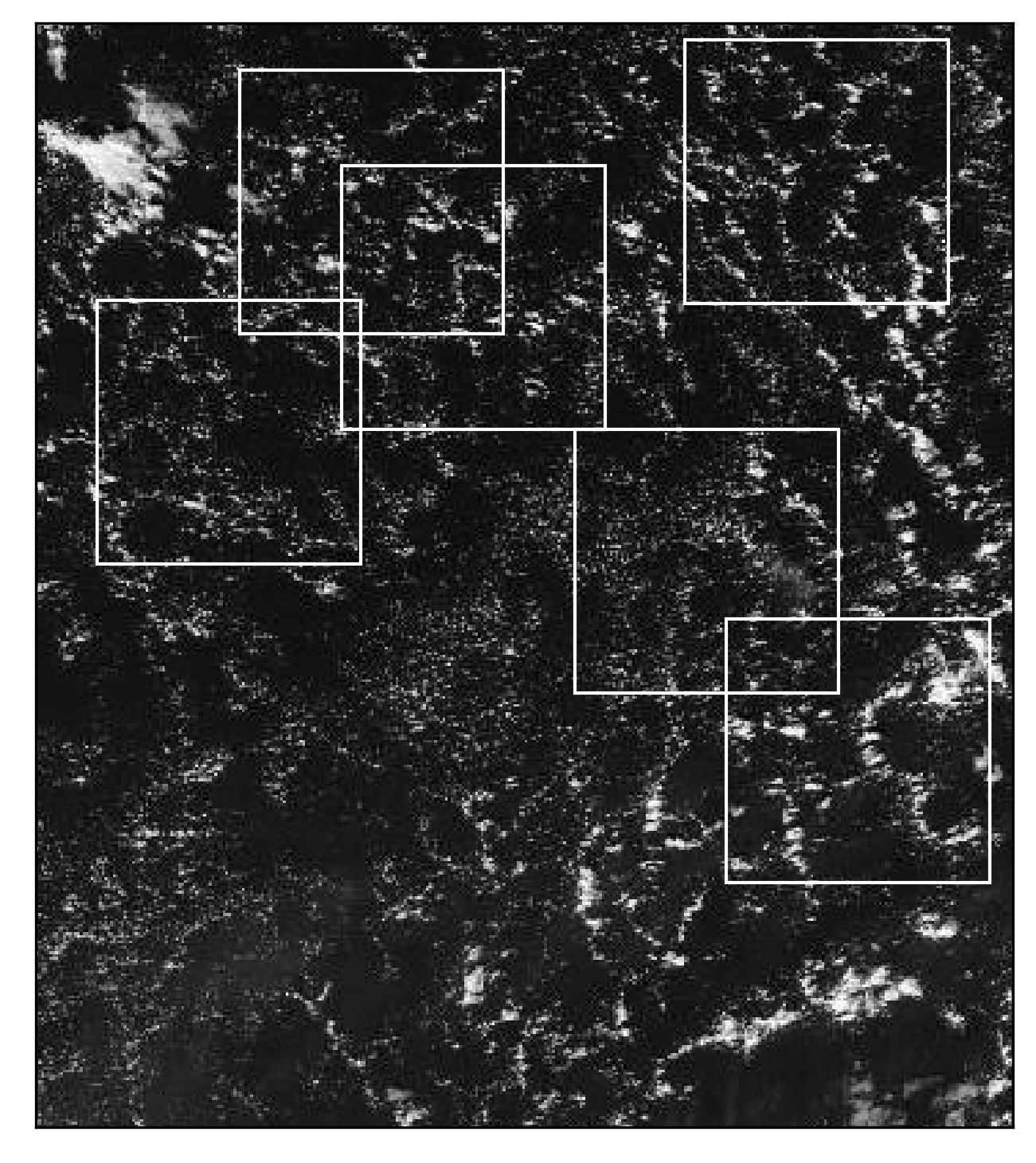}
        \caption{Most extreme sugar example.}
        \label{subfig:extreme_sugar_ex}
    \end{subfigure}
    \hfill
    \begin{subfigure}{0.3\textwidth}
        \centering
        \includegraphics[width=\linewidth]{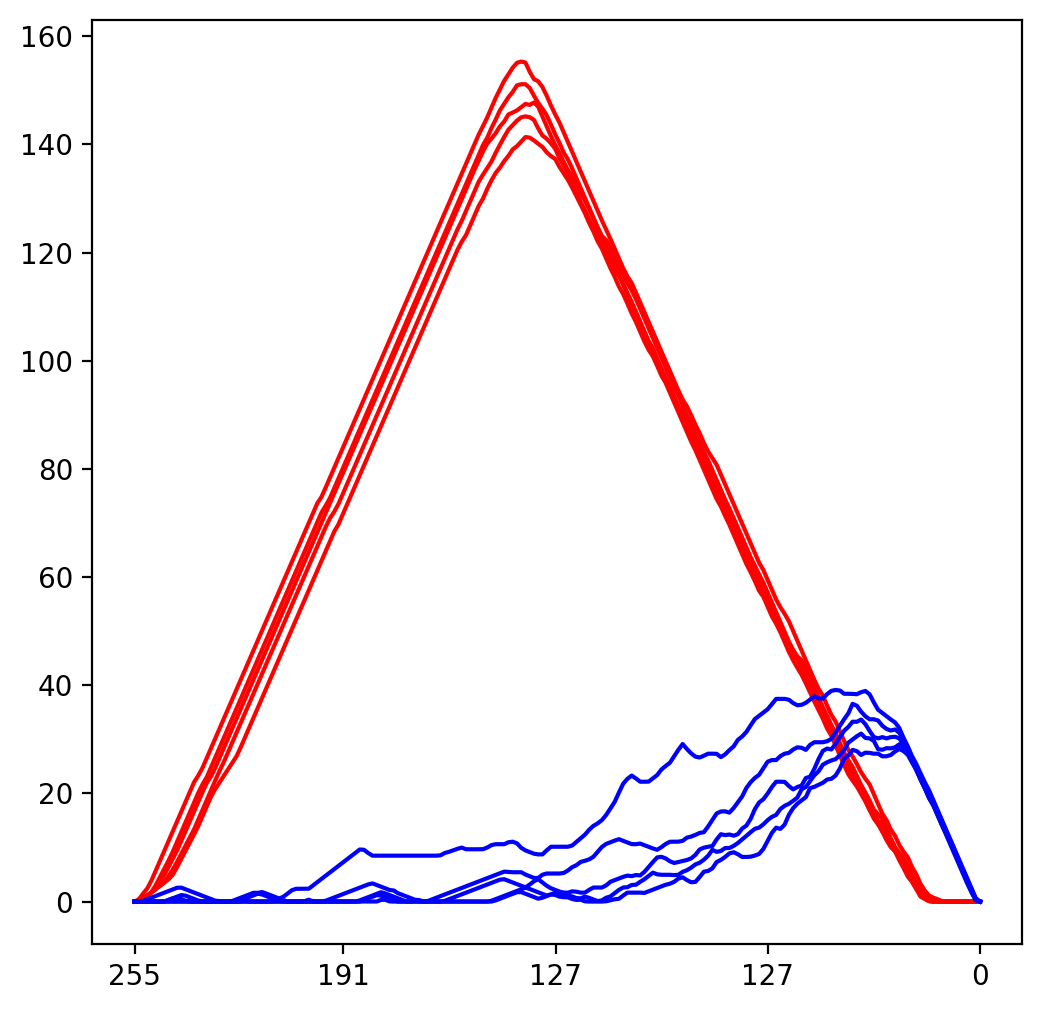}
        \caption{Landscape for the sample in \ref{subfig:extreme_sugar_ex}.}
        \label{subfig:extreme_sugar_lscape}
    \end{subfigure}
    \hfill
    \begin{subfigure}{0.3\textwidth}
        \centering
        \includegraphics[width=\linewidth]{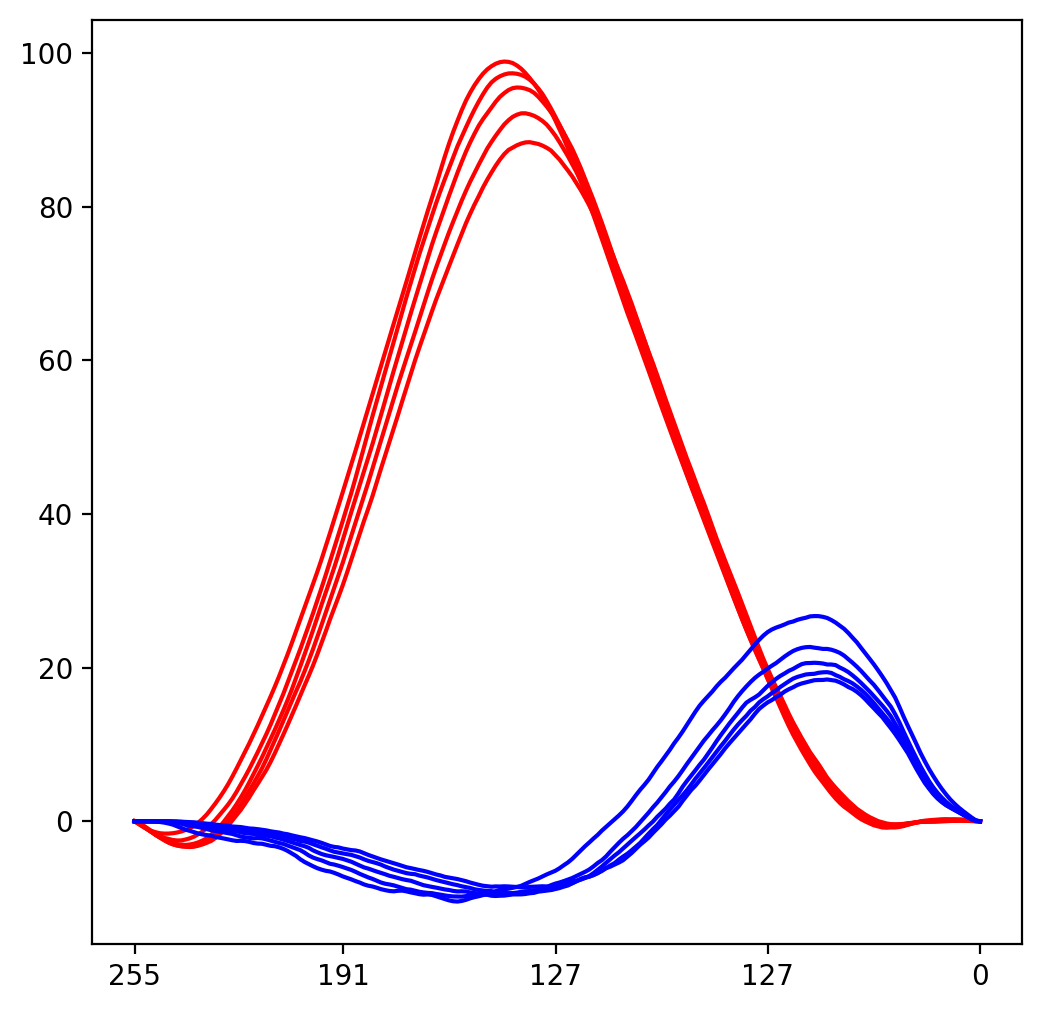}
        \caption{``Virtual'' sugar landscape.}
        \label{subfig:extreme_sugar_virtual}
    \end{subfigure}

    \medskip

    \begin{subfigure}{0.3\textwidth}
        \centering
        \includegraphics[width=\linewidth]{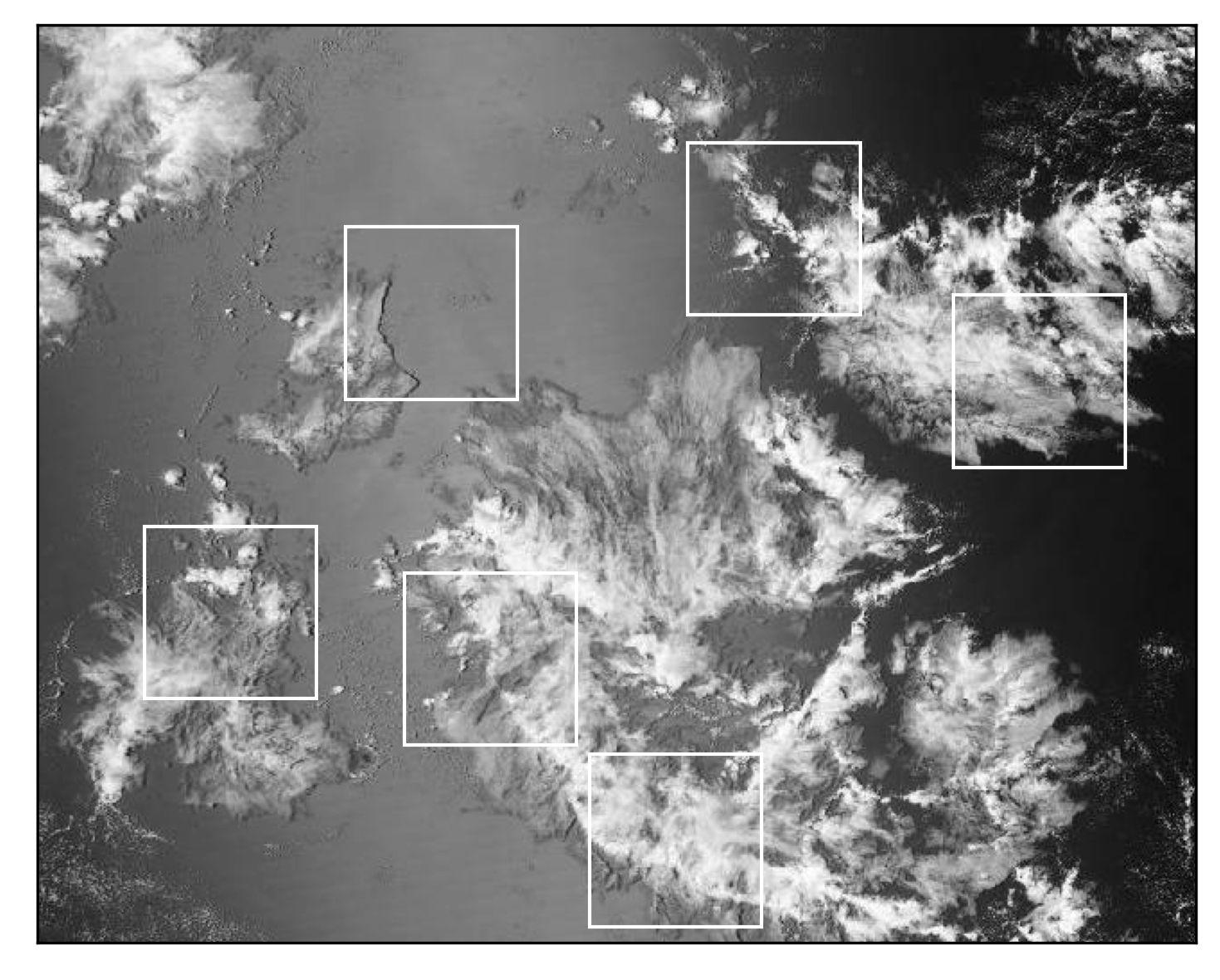}
        \caption{Most extreme flowers example.}
        \label{subfig:extreme_flower_ex}
    \end{subfigure}
    \hfill
    \begin{subfigure}{0.3\textwidth}
        \centering
        \includegraphics[width=\linewidth]{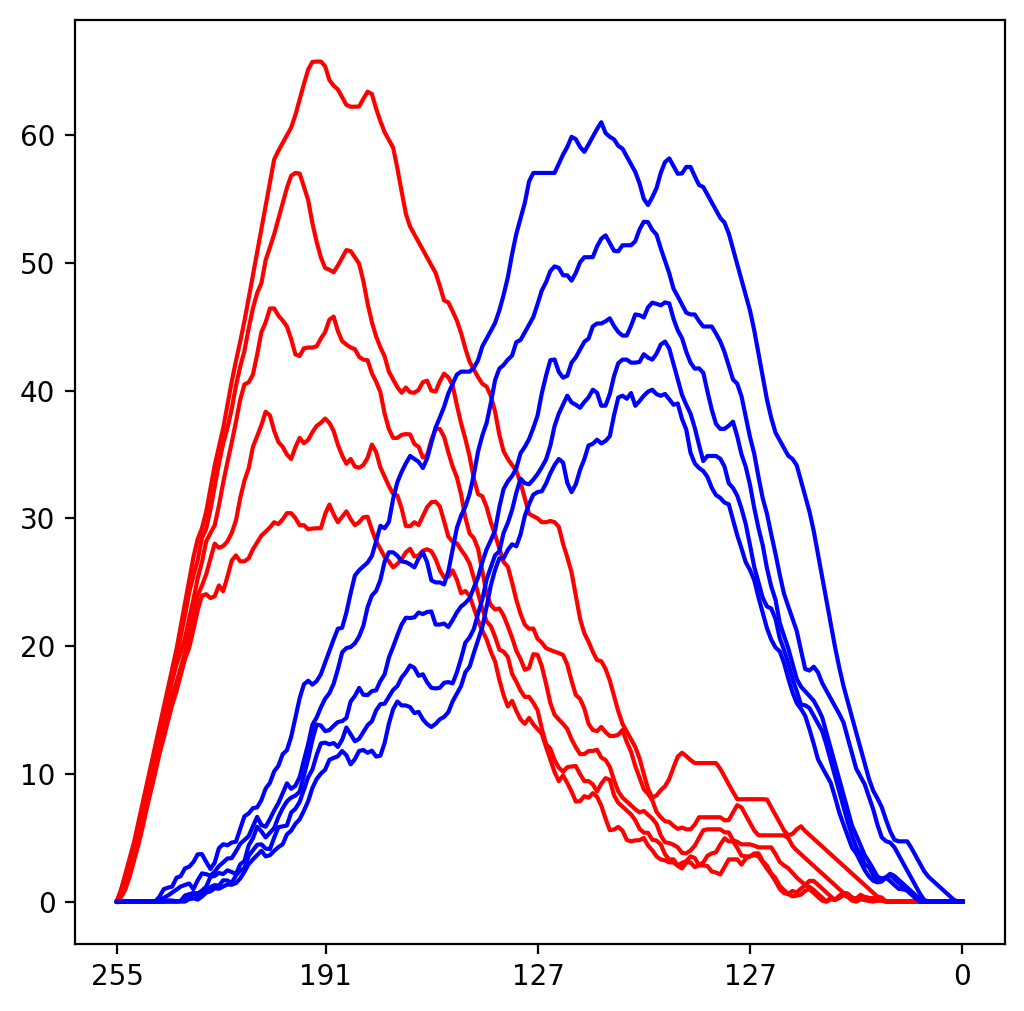}
        \caption{Landscape for the sample in \ref{subfig:extreme_flower_ex}.}
        \label{subfig:extreme_flower_lscape}
    \end{subfigure}
    \hfill
    \begin{subfigure}{0.3\textwidth}
        \centering
        \includegraphics[width=\linewidth]{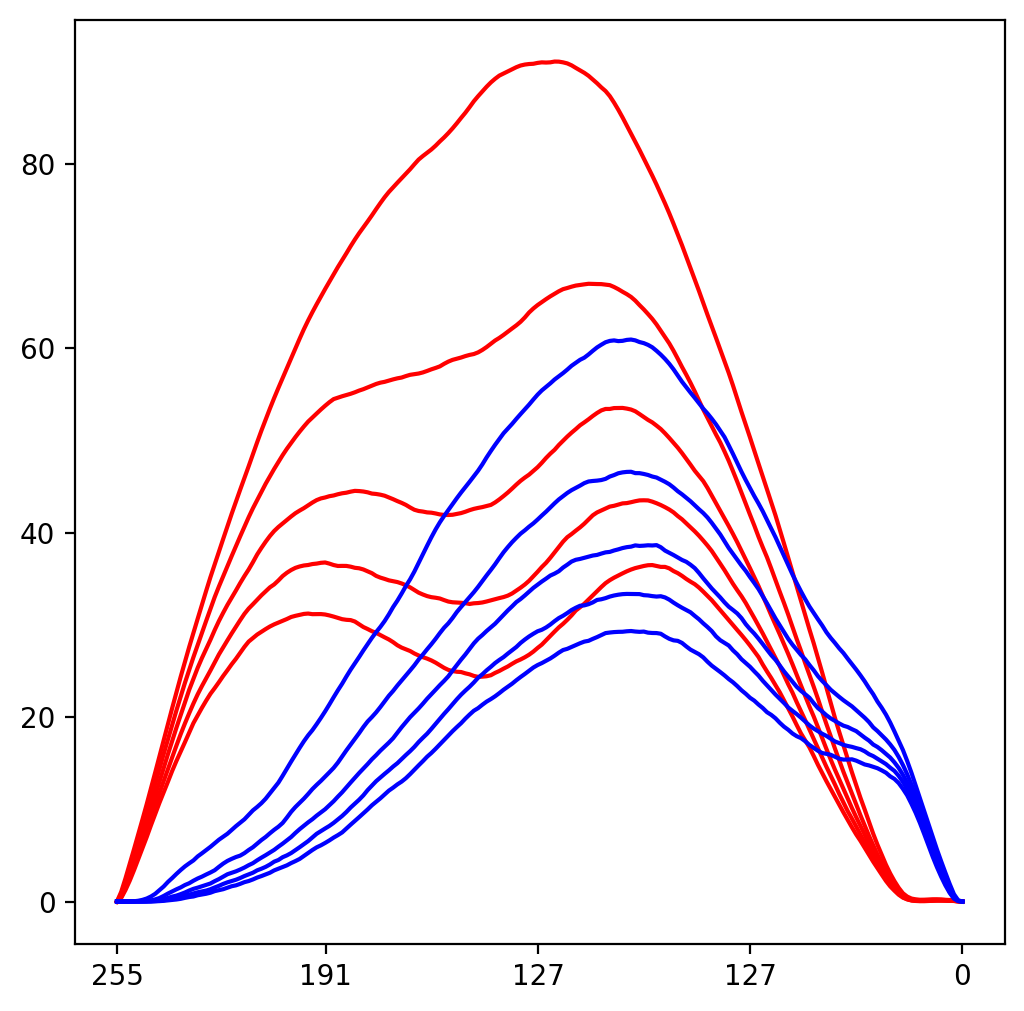}
        \caption{``Virtual'' flowers landscape.}
        \label{subfig:extreme_flower_virtual}
    \end{subfigure}

    \caption{Samples showing the sugar and flowers samples farthest from the separating hyperplane (in \ref{subfig:extreme_sugar_ex} and \ref{subfig:extreme_flower_ex}), and their landscapes (in \ref{subfig:extreme_sugar_lscape} and \ref{subfig:extreme_sugar_lscape}).
    The ``virtual'' landscapes obtained by traveling along the line normal to the separating hyperplane are shown in \ref{subfig:extreme_sugar_virtual} and \ref{subfig:extreme_flower_virtual}.}
    \label{fig:extreme_examples}
\end{figure}

To begin, we explore what can be learned just from the summarized data, without looking at examples.
To discover what the separating plane between sugar points and flowers points represents, we create ``virtual'' landscapes.
We first lift the separating plane in $\R^3$ to the hyperplane in $\R^{2000}$ consisting of all the points that project (under PCA) into the separating plane in $\R^3$.
Next, we find the line normal to this hyperplane that passes through geometric center of the data.
Finally, we choose points on this line that fall at the outer extent of the data point cloud.
These points live in the landscape embedding space ($\R^{2000}$), but are not sampled data points.
However, by applying the inverse landscape embedding, we can visualize the landscape-like set of curves that would give this embedded point.
Virtual landscapes for sugar and flowers can be seen in Figures~\ref{subfig:extreme_sugar_virtual} and~\ref{subfig:extreme_flower_virtual}, respectively.

An advantage of this approach is that it synthesizes trends from the real data into a readable, controllable format that demonstrates how SVM is separating these classes.
When we compare these virtual landscapes with the actual landscapes farthest from the separating hyperplane (seen in Figures~\ref{subfig:extreme_sugar_lscape} and~\ref{subfig:extreme_flower_lscape}), we can see that the virtual landscapes are smoother, but that the overall shapes are remarkably similar.

We can also interpret the shapes of these landscapes in terms of the features present in the images.
Let us examine the images and corresponding landscapes in Figure~\ref{fig:extreme_examples}.
The most prominent feature in the two landscapes is the tall red peak in the sugar landscape, shown in Figure~\ref{subfig:extreme_sugar_lscape}.
Recall that the red lines denote 0-dimensional homology (connected components) while the blue lines denote $1$-dimensional homology (holes).
This red peak represents the presence of strongly persistent connected components, i.e., separated regions of bright cloud strongly contrasting against a much darker background.
The sharpness of this peak also indicates that these features are similar in both the intensity of the cloud top and the intensity of the surrounding background.
The comparatively low blue curves with only a small peak at the end indicate a lack of 1-dimensional homological features (holes), and thus the texture within connected components is relatively uniform.
Looking at the image in Figure~\ref{subfig:extreme_sugar_ex}, we see these observations borne out: there are numerous small clouds of similar brightnesses which stand in stark contrast to the overall uniformly dark background, matching the tall $H_0$ peak.
Because these clouds are relatively small, there is little discernible texture within each cloud, which corresponds to the relative absence of $H_1$ features.

On the other hand, the flowers landscape in Figure~\ref{subfig:extreme_flower_lscape} displays a lower red peak, with more separated curves.
This lack of concentration indicates that there is more variation in the intensity values at which connected components appear (the brightest part of the component) and at which they merge together (the intensity of the bridge connecting that component to another), while the lower height indicates that these features are overall less persistent -- they merge into one another more rapidly.
Additionally, there is a much stronger $H_1$ signal in this case than in the sugar landscape, meaning that the connected components have more internal texture, with numerous holes appearing and disappearing over a wide range of intensity values.
These observations match with what we see in Figure~\ref{subfig:extreme_flower_ex}.
The clouds in this image are much larger and cover more of the frame, with varying intensities within and between the clouds, leading to the more varied $H_0$ landscape.
This image also shows much more internal texture to the clouds, with far more of a dimpling effect than in the sugar example.

In summary, this example shows how the patterns learned by the TDA-SVM algorithm can be translated back to homological features which in turn correspond to weather-relevant features in the original image.
This is made possible by the fact that the SVM model can be represented by a single separating plane, which can be translated back into the space of persistence landscapes and then interpreted, yielding a highly interpretable approach to the pairwise classification problem.

\subsection{Comparison of this classifier to those in \citet{rasp2020combining}} 
\label{subsec:comparison}

{\bf Accuracy:}
The accuracy of our approach cannot be directly compared to the deep learning algorithms in \citet{rasp2020combining} because they address different tasks.
The task considered here is to (1) choose a single class (out of two) for an annotation assumed to consist of a single cloud type, 
(2) based on several small patches ($96 \times 96$ pixels).
In contrast, the task considered in \citet{rasp2020combining} is much more complex, namely to (1) assign one or more labels for an annotation; (2) based on a very large image.
We choose the simpler task for our TDA approach in order to expose the properties of a TDA algorithm, and trying to implement a multi-label assignment (for example by using a sliding window approach) likely would have made this exploration more complicated without providing new insights.
However, even without a direct comparison it is obvious from the results that this first TDA-SVM approach cannot nearly achieve the accuracy of the deep learning approaches.

{\bf Required data samples:} Our approach only requires a few hundred labeled data samples to develop a classifier.
This reduces the required labeling effort by two orders of magnitude relative to the tens of thousands of labeled samples in \citet{rasp2020combining}.

{\bf Computational effort:}  
Computations were performed on a Surface Pro 6 with an Intel Core i5-8250U CPU.
The computational bottleneck in this case was computing the persistent homology -- for 800 samples (and therefore 4800 subsamples to compute persistent homology for), approximately 45 minutes of wall-clock time was required.
This is already much less computational time than is generally required to train a deep network, and it is likely that this could be significantly improved by parallelizing, as each sample can be processed entirely separately.

{\bf Interpretability and failure modes:} Our approach yields a highly interpretable model that provides an intuitive explanation of how the algorithm distinguishes different classes, while the deep learning methods do not.
Furthermore, the interpretation of the separation plane in our model makes it easy to provide insights into failure modes, i.e.\ which types of mesoscale patterns can be easily or not so easily be distinguished by their topological features, and thus by this approach.

\section{Advanced TDA concepts}
\label{sec:advanced}

In this section we briefly discuss and provide references for some advanced TDA concepts that are beyond the scope of this article, along with motivations for when and why readers might find them useful.

Figure~\ref{fig:homeomorphism_ex} shows that while persistent homology measures some spatial aspects of the intensity function, it is also invariant under nice deformations (``homeomorphsims'') of the domain.
However, there is another (very popular) type of persistent homology, constructed using growing offsets of a shape, or unions of growing balls, that distinguishes between different deformations of the domain~\citep{carlsson2009topology,ghrist2008barcodes}.
We expect that this variant of persistent homology will also find applications in atmospheric science, and we refer the reader to~\citet{tymochko2020using} for such an example.

We are particularly interested to explore the use of TDA to analyze cloud properties from satellite imagery, e.g., to detect convection.
While the example here looked at large scale organization of clouds, to analyze properties like convection we would zoom far into a single cloud and analyze its texture, e.g., seek to identify whether there is a ``bubbling'' texture apparent in a considered area of the cloud.
Preliminary analysis leads us to believe that it might be necessary to use more sophisticated TDA tools for this purpose than discussed here, such as {\it vineyards}~\citep{cohen2006vines} or {\it crocker plots}~\citep{topaz2015topological}, which incorporate temporal context by analyzing the topological properties of {\it sequences} of images, rather than individual images.

We have considered persistent homology that varies over a single parameter -- the intensity of the satellite image.
However, one frequently encounters situations in which two or more parameters naturally arise.
For example, one can perform superlevelset persistent homology on a 2-channel image, containing the intensities with respect to two frequencies, with respect to the parameter from either the first channel or the second.
In these contexts, multiparameter persistence~\citep{carlsson2009computing,carlsson2009theory,cerri2013betti,lesnick2016rivet} allows one to consider both parameters at once, even though the underlying mathematics is more subtle and computations are more difficult.
A version of multiparameter persistence was applied recently to the atmospheric domain in \citet{strommen2021topological}.

Persistence barcodes and diagrams are not ideal as inputs into machine learning algorithms, because they are not vectors residing in a linear space.
This is evidenced by the fact that averages of persistence diagrams need not be unique~\citep{mileyko2011probability}.
There are a wide array of options for transforming persistence diagrams for use in machine learning, including not only persistence landscapes~\citep{bubenik2015statistical} but also persistence images~\citep{adams2017persistence} and stable kernels~\citep{reininghaus2015stable}, for example.
TDA has been gaining traction in machine learning tasks as more tools become available to integrate it into existing workflows, in both neural network layers~\citep{moroni2021learning} and loss functions~\citep{clough2020topological}.
As an example application, TDA has recently been used to compare models with differing grids and resolutions~\citep{oforiboateng2021application}.

There is a variant of superlevelset persistent homology, called \emph{extended} persistent homology~\citep[Section~VII.3]{edelsbrunner2010computational}, which performs two sweeps (instead of just one) over the range of intensity values.
Extended superlevelset persistent homology detects all of the features measured by superlevelset persistent homology, plus more.
It may be the case that one can extract more discriminative information from a satellite image by instead computing the extended persistence diagram.

\section{Conclusions and Future Work} 
\label{sec:conclusion}

The primary contributions of this manuscript are as follows.
(1) This paper presents, to the best of our knowledge, the first attempt to provide a comprehensive, easy-to-understand introduction to popular TDA concepts customized for the environmental science community.
In particular, we seek to provide readers with an intuitive understanding of the topological properties that can be extracted using TDA by translating cloud imagery into persistence landscapes, interpreting the landscapes, then highlighting the topological properties in the original images.
(2) In a case study, we demonstrate step-by-step the process of applying TDA, combined with a simple machine learning model (SVM), to extract information from real-world meteorological imagery.
The case study focuses on how to use TDA to classify mesoscale organization of clouds from satellite imagery, which has never been addressed by TDA before.
(3) The most novel contribution is the  interpretation procedure we developed that projects the class separation planes identified by the SVM algorithm back into topological space.
This in turns allows us to fully understand the strategy used by the classifier in meteorological image space, thus providing a fully interpretable classifier.

In future work we seek to explore several of the advanced methods outlined in Section \ref{sec:advanced}.
We believe that there are many applications to be explored with TDA, including the applications suggested by 
\citet{rasp2020combining} for their methods, namely 
{\it ``detecting atmospheric rivers and tropical cyclones in satellite and model output, classifying ice and snow particles images obtained from cloud probe imagery, or even large-scale weather regimes''}~\citep{rasp2020combining}.
Furthermore, 
as discussed in Section~\ref{sec:intro}, TDA has already been shown to be useful to identify certain properties of atmospheric rivers, wildfires, and hurricanes, and we expect TDA to find additional use in those areas as well.
Our group is particularly interested in using TDA to detect convection in clouds, and to distinguish blowing dust from, say, blowing snow, in satellite imagery.

We have only scratched the surface here of exploring how TDA can support image analysis tasks in environmental science, but we hope that this primer will accelerate the use of TDA for this purpose.


\section*{Acknowledgements}

The work by the first and last authors was supported in part by the National Science Foundation under Grant No.\ OAC-1934668, and under Grant No.\ ICER-2019758 which funds the NSF AI Institute for Research on Trustworthy AI in Weather, Climate, and Coastal Oceanography.
The second author thanks the Institute of Science and Technology Austria for hosting him during this research.
We thank Charles H. White at CIRA (CSU) for constructive feedback and excellent suggestions for this manuscript.
The authors would also like to thank \citet{rasp2020combining} for making their dataset publicly available.

\section*{Data Availability Statement}
The underlying crowd-sourced data from \citet{rasp2020combining} are available at https://github.com/raspstephan/sugar-flower-fish-or-gravel.
The code used in our analysis is available in a GitHub repository at \url{https://github.com/zyjux/sffg_tda}.

\bibliographystyle{ametsocV6}
\bibliography{sffg_paper}

\begin{thebibliography}{53}
\providecommand{\natexlab}[1]{#1}
\providecommand{\url}[1]{\texttt{#1}}
\renewcommand{\UrlFont}{\rmfamily}
\providecommand{\urlprefix}{URL }
\expandafter\ifx\csname urlstyle\endcsname\relax
  \providecommand{\doi}[1]{https://doi.org/\discretionary{}{}{}#1}\else
  \providecommand{\doi}{https://doi.org/\discretionary{}{}{}\begingroup
  \urlstyle{rm}\Url}\fi
\providecommand{\eprint}[2][]{\url{#2}}

\bibitem[{Adams et~al.(2017)}]{adams2017persistence}
Adams, H., and Coauthors, 2017: Persistence images: {A} vector representation
  of persistent homology. \textit{Journal of Machine Learning Research},
  \textbf{18~(8)}, 1--35.

\bibitem[{Boser et~al.(1992)Boser, Guyon,, and Vapnik}]{boser1992training}
Boser, B.~E., I.~M. Guyon, and V.~N. Vapnik, 1992: A training algorithm for
  optimal margin classifiers. \textit{Proceedings of the fifth annual workshop
  on Computational learning theory - {COLT} {\textquotesingle}92}, {ACM} Press,
  \doi{10.1145/130385.130401}.

\bibitem[{Brenowitz et~al.(2020)Brenowitz, Beucler, Pritchard,, and
  Bretherton}]{brenowitz2020interpreting}
Brenowitz, N.~D., T.~Beucler, M.~Pritchard, and C.~S. Bretherton, 2020:
  Interpreting and stabilizing machine-learning parametrizations of convection.
  \textit{Journal of the Atmospheric Sciences}, \textbf{77~(12)}, 4357--4375.

\bibitem[{Bubenik(2015)}]{bubenik2015statistical}
Bubenik, P., 2015: Statistical topological data analysis using persistence
  landscapes. \textit{J. Mach. Learn. Res.}, \textbf{16~(1)}, 77--102.

\bibitem[{Carlsson(2009)}]{carlsson2009topology}
Carlsson, G., 2009: Topology and data. \textit{Bulletin of the American
  Mathematical Society}, \textbf{46~(2)}, 255--308.

\bibitem[{Carlsson et~al.(2009)Carlsson, Singh,, and
  Zomorodian}]{carlsson2009computing}
Carlsson, G., G.~Singh, and A.~Zomorodian, 2009: Computing multidimensional
  persistence. \textit{International Symposium on Algorithms and Computation},
  Springer, 730--739.

\bibitem[{Carlsson and Zomorodian(2009)Carlsson, and
  Zomorodian}]{carlsson2009theory}
Carlsson, G., and A.~Zomorodian, 2009: The theory of multidimensional
  persistence. \textit{Discrete \& Computational Geometry}, \textbf{42~(1)},
  71--93.

\bibitem[{Cerri et~al.(2013)Cerri, Fabio, Ferri, Frosini,, and
  Landi}]{cerri2013betti}
Cerri, A., B.~D. Fabio, M.~Ferri, P.~Frosini, and C.~Landi, 2013: Betti numbers
  in multidimensional persistent homology are stable functions.
  \textit{Mathematical Methods in the Applied Sciences}, \textbf{36~(12)},
  1543--1557.

\bibitem[{Chung et~al.(2009)Chung, Bubenik,, and Kim}]{chung2009persistence}
Chung, M.~K., P.~Bubenik, and P.~T. Kim, 2009: Persistence diagrams of cortical
  surface data. \textit{International Conference on Information Processing in
  Medical Imaging}, Springer, 386--397.

\bibitem[{Clough et~al.(2020)Clough, Byrne, Oksuz, Zimmer, Schnabel,, and
  King}]{clough2020topological}
Clough, J., N.~Byrne, I.~Oksuz, V.~A. Zimmer, J.~A. Schnabel, and A.~King,
  2020: A topological loss function for deep-learning based image segmentation
  using persistent homology. \textit{IEEE Transactions on Pattern Analysis and
  Machine Intelligence}, 1--1, \doi{10.1109/TPAMI.2020.3013679}.

\bibitem[{Cohen-Steiner et~al.(2006)Cohen-Steiner, Edelsbrunner,, and
  Morozov}]{cohen2006vines}
Cohen-Steiner, D., H.~Edelsbrunner, and D.~Morozov, 2006: Vines and vineyards
  by updating persistence in linear time. \textit{Proceedings of the
  twenty-second annual symposium on Computational geometry}, ACM, 119--126.

\bibitem[{Denby(2020)}]{denby2020discovering}
Denby, L., 2020: Discovering the importance of mesoscale cloud organization
  through unsupervised classification. \textit{Geophysical Research Letters},
  \textbf{47~(1)}, e2019GL085\,190.

\bibitem[{Deshmukh et~al.(2022)Deshmukh, Baskar, Bradley, Berger,, and
  Meiss}]{deshmukh2022machine}
Deshmukh, V., S.~Baskar, E.~Bradley, T.~Berger, and J.~D. Meiss, 2022: Machine
  learning approaches to solar-flare forecasting: {I}s complex better?
  \textit{arXiv preprint arXiv:2202.08776}.

\bibitem[{Ebert-Uphoff and Hilburn(2020)Ebert-Uphoff, and
  Hilburn}]{ebert2020evaluation}
Ebert-Uphoff, I., and K.~Hilburn, 2020: Evaluation, tuning, and interpretation
  of neural networks for working with images in meteorological applications.
  \textit{Bulletin of the American Meteorological Society}, \textbf{101~(12)},
  E2149--E2170.

\bibitem[{Edelsbrunner and Harer(2010)Edelsbrunner, and
  Harer}]{edelsbrunner2010computational}
Edelsbrunner, H., and J.~L. Harer, 2010: \textit{Computational Topology: An
  Introduction}. American Mathematical Society, Providence.

\bibitem[{Fletcher and Islam(2018)Fletcher, and Islam}]{fletcher2018comparing}
Fletcher, S., and M.~Z. Islam, 2018: Comparing sets of patterns with the
  {J}accard index. \textit{Australasian Journal of Information Systems},
  \textbf{22}.

\bibitem[{Gagne~II et~al.(2019)Gagne~II, Haupt, Nychka,, and
  Thompson}]{gagne2019interpretable}
Gagne~II, D.~J., S.~E. Haupt, D.~W. Nychka, and G.~Thompson, 2019:
  Interpretable deep learning for spatial analysis of severe hailstorms.
  \textit{Monthly Weather Review}, \textbf{147~(8)}, 2827--2845.

\bibitem[{Gardner et~al.(2022)Gardner, Hermansen, Pachitariu, Burak, Baas,
  Dunn, Moser,, and Moser}]{Gardner2022Nature}
Gardner, R.~J., E.~Hermansen, M.~Pachitariu, Y.~Burak, N.~A. Baas, B.~A. Dunn,
  M.-B. Moser, and E.~I. Moser, 2022: Toroidal topology of population activity
  in grid cells. \textit{Nature}, \textbf{602}, 123--128,
  \doi{10.1038/s41586-021-04268-7}.

\bibitem[{Gentine et~al.(2018)Gentine, Pritchard, Rasp, Reinaudi,, and
  Yacalis}]{gentine2018could}
Gentine, P., M.~Pritchard, S.~Rasp, G.~Reinaudi, and G.~Yacalis, 2018: Could
  machine learning break the convection parameterization deadlock?
  \textit{Geophysical Research Letters}, \textbf{45~(11)}, 5742--5751.

\bibitem[{Ghrist(2008)}]{ghrist2008barcodes}
Ghrist, R., 2008: Barcodes: {T}he persistent topology of data. \textit{Bulletin
  of the American Mathematical Society}, \textbf{45~(1)}, 61--75.

\bibitem[{Gumley et~al.(2010)Gumley, Descloitres,, and
  Schmaltz}]{gumley2010creating}
Gumley, L., J.~Descloitres, and J.~Schmaltz, 2010: Creating reprojected true
  color {MODIS} images: A tutorial. Tech. rep., Space Science and Engineering
  Center.
  \urlprefix\url{https://cdn.earthdata.nasa.gov/conduit/upload/946/MODIS_True_Color.pdf}.

\bibitem[{ITU-R(2011)}]{itur2011}
ITU-R, 2011: Studio encoding parameters of digital television for standard 4:3
  and wide-screen 16:9 aspect ratios. Tech. Rep. BT.601, International
  Telecommunication Union.
  \urlprefix\url{https://www.itu.int/dms_pubrec/itu-r/rec/bt/R-REC-BT.601-7-201103-I!!PDF-E.pdf}.

\bibitem[{Jaccard(1901)}]{jaccard1901etude}
Jaccard, P., 1901: {\'E}tude comparative de la distribution florale dans une
  portion des alpes et des jura. \textit{Bull Soc Vaudoise Sci Nat},
  \textbf{37}, 547--579.

\bibitem[{Kim and Vogel(2019)Kim, and Vogel}]{kim2019deciphering}
Kim, H., and C.~Vogel, 2019: Deciphering active wildfires in the {S}outhwestern
  {USA} using topological data analysis. \textit{Climate}, \textbf{7~(12)},
  135.

\bibitem[{Kram{\'a}r et~al.(2016)Kram{\'a}r, Levanger, Tithof, Suri, Xu, Paul,
  Schatz,, and Mischaikow}]{kramar2016analysis}
Kram{\'a}r, M., R.~Levanger, J.~Tithof, B.~Suri, M.~Xu, M.~Paul, M.~F. Schatz,
  and K.~Mischaikow, 2016: Analysis of {K}olmogorov flow and
  {R}ayleigh--{B}{\'e}nard convection using persistent homology.
  \textit{Physica D: Nonlinear Phenomena}, \textbf{334}, 82--98.

\bibitem[{Krasnopolsky et~al.(2005)Krasnopolsky, Fox-Rabinovitz,, and
  Chalikov}]{krasnopolsky2005new}
Krasnopolsky, V.~M., M.~S. Fox-Rabinovitz, and D.~V. Chalikov, 2005: New
  approach to calculation of atmospheric model physics: Accurate and fast
  neural network emulation of longwave radiation in a climate model.
  \textit{Monthly Weather Review}, \textbf{133~(5)}, 1370--1383.

\bibitem[{Lawson et~al.(2019)Lawson, Sholl, Brown, Fasy,, and
  Wenk}]{lawson2019persistent}
Lawson, P., A.~B. Sholl, J.~Brown, B.~T. Fasy, and C.~Wenk, 2019: Persistent
  homology for the quantitative evaluation of architectural features in
  prostate cancer histology. \textit{Scientific reports}, \textbf{9~(1)},
  1--15.

\bibitem[{L'Ecuyer et~al.(2015)}]{lecuyer2015observed}
L'Ecuyer, T.~S., and Coauthors, 2015: The observed state of the energy budget
  in the early twenty-first century. \textit{Journal of Climate},
  \textbf{28~(21)}, 8319--8346, \doi{10.1175/jcli-d-14-00556.1}.

\bibitem[{Lesnick and Wright(2016)Lesnick, and Wright}]{lesnick2016rivet}
Lesnick, M., and M.~Wright, 2016: {RIVET}: The rank invariant visualization and
  exploration tool. \textit{Software available at
  \url{https://github.com/rivetTDA}}.

\bibitem[{McGovern et~al.(2022)McGovern, Ebert-Uphoff, Gagne,, and
  Bostrom}]{mcgovern2022we}
McGovern, A., I.~Ebert-Uphoff, D.~J. Gagne, and A.~Bostrom, 2022: Why we need
  to focus on developing ethical, responsible, and trustworthy artificial
  intelligence approaches for environmental science. \textit{Environmental Data
  Science}, \textbf{1}.

\bibitem[{McGovern et~al.(2019)McGovern, Lagerquist, Gagne, Jergensen, Elmore,
  Homeyer,, and Smith}]{mcgovern2019making}
McGovern, A., R.~Lagerquist, D.~J. Gagne, G.~E. Jergensen, K.~L. Elmore, C.~R.
  Homeyer, and T.~Smith, 2019: Making the black box more transparent:
  Understanding the physical implications of machine learning. \textit{Bulletin
  of the American Meteorological Society}, \textbf{100~(11)}, 2175--2199.

\bibitem[{Merritt and Seymour(2021)Merritt, and
  Seymour}]{merritt2021visualizing}
Merritt, R.~B., and L.~Seymour, 2021: Visualizing planetary rossby waves with
  topological data analysis. M.S. thesis, University of Georgia, 54 pp., Ann
  Arbor,
  \urlprefix\url{https://esploro.libs.uga.edu/esploro/outputs/9949375354302959}.

\bibitem[{Mileyko et~al.(2011)Mileyko, Mukherjee,, and
  Harer}]{mileyko2011probability}
Mileyko, Y., S.~Mukherjee, and J.~Harer, 2011: Probability measures on the
  space of persistence diagrams. \textit{Inverse Problems}, \textbf{27~(12)},
  124\,007.

\bibitem[{Moroni and Pascali(2021)Moroni, and Pascali}]{moroni2021learning}
Moroni, D., and M.~A. Pascali, 2021: Learning topology: Bridging computational
  topology and machine learning. \textit{Pattern Recognition and Image
  Analysis}, \textbf{31~(3)}, 443--453, \doi{10.1134/s1054661821030184}.

\bibitem[{Muszynski et~al.(2019)Muszynski, Kashinath, Kurlin,, and
  Wehner}]{muszynski2019topological}
Muszynski, G., K.~Kashinath, V.~Kurlin, and M.~Wehner, 2019: Topological data
  analysis and machine learning for recognizing atmospheric river patterns in
  large climate datasets. \textit{Geoscientific Model Development},
  \textbf{12~(2)}, 613--628.

\bibitem[{Ofori-Boateng et~al.(2021)Ofori-Boateng, Lee, Gorski, Garay,, and
  Gel}]{oforiboateng2021application}
Ofori-Boateng, D., H.~Lee, K.~M. Gorski, M.~J. Garay, and Y.~R. Gel, 2021:
  Application of topological data analysis to multi-resolution matching of
  aerosol optical depth maps. \textit{Frontiers in Environmental Science},
  \textbf{9}, \doi{10.3389/fenvs.2021.684716}.

\bibitem[{Rasp et~al.(2018)Rasp, Pritchard,, and Gentine}]{rasp2018deep}
Rasp, S., M.~S. Pritchard, and P.~Gentine, 2018: Deep learning to represent
  subgrid processes in climate models. \textit{Proceedings of the National
  Academy of Sciences}, \textbf{115~(39)}, 9684--9689.

\bibitem[{Rasp et~al.(2020)Rasp, Schulz, Bony,, and
  Stevens}]{rasp2020combining}
Rasp, S., H.~Schulz, S.~Bony, and B.~Stevens, 2020: Combining crowdsourcing and
  deep learning to explore the mesoscale organization of shallow convection.
  \textit{Bulletin of the American Meteorological Society}, \textbf{101~(11)},
  E1980--E1995.

\bibitem[{Reininghaus et~al.(2015)Reininghaus, Huber, Bauer,, and
  Kwitt}]{reininghaus2015stable}
Reininghaus, J., S.~Huber, U.~Bauer, and R.~Kwitt, 2015: A stable multi-scale
  kernel for topological machine learning. \textit{Proceedings of the IEEE
  Conference on Computer Vision and Pattern Recognition}, 4741--4748.

\bibitem[{Rudin(2019)}]{rudin2019stop}
Rudin, C., 2019: Stop explaining black box machine learning models for high
  stakes decisions and use interpretable models instead. \textit{Nature Machine
  Intelligence}, \textbf{1~(5)}, 206--215.

\bibitem[{Schmit et~al.(2017)Schmit, Griffith, Gunshor, Daniels, Goodman,, and
  Lebair}]{schmit2017closer}
Schmit, T.~J., P.~Griffith, M.~M. Gunshor, J.~M. Daniels, S.~J. Goodman, and
  W.~J. Lebair, 2017: A closer look at the {ABI} on the {GOES}-r series.
  \textit{Bulletin of the American Meteorological Society}, \textbf{98~(4)},
  681--698, \doi{10.1175/bams-d-15-00230.1}.

\bibitem[{Schultz et~al.(2021)Schultz, Betancourt, Gong, Kleinert, Langguth,
  Leufen, Mozaffari,, and Stadtler}]{schultz2021can}
Schultz, M., C.~Betancourt, B.~Gong, F.~Kleinert, M.~Langguth, L.~Leufen,
  A.~Mozaffari, and S.~Stadtler, 2021: Can deep learning beat numerical weather
  prediction? \textit{Philosophical Transactions of the Royal Society A},
  \textbf{379~(2194)}, 20200\,097.

\bibitem[{Schwartz et~al.(2020)Schwartz, Dodge, Smith,, and
  Etzioni}]{schwartz2020green}
Schwartz, R., J.~Dodge, N.~A. Smith, and O.~Etzioni, 2020: Green {AI}.
  \textit{Communications of the ACM}, \textbf{63~(12)}, 54--63.

\bibitem[{Segovia-Dominguez et~al.(2021)Segovia-Dominguez, Zhen, Wagh, Lee,,
  and Gel}]{segoviadominguez2021tlifelstm}
Segovia-Dominguez, I., Z.~Zhen, R.~Wagh, H.~Lee, and Y.~R. Gel, 2021:
  {TLife}-{LSTM}: Forecasting future {COVID}-19 progression with topological
  signatures of atmospheric conditions. \textit{Advances in Knowledge Discovery
  and Data Mining}, Springer International Publishing, 201--212,
  \doi{10.1007/978-3-030-75762-5_17}.

\bibitem[{Sena et~al.(2021)Sena, da~Paix{\~{a}}o,, and
  de~Almeida~Fran{\c{c}}a}]{sena2021topological}
Sena, C. {\'{A}}.~P., J.~A.~R. da~Paix{\~{a}}o, and J.~R.
  de~Almeida~Fran{\c{c}}a, 2021: A topological data analysis approach for
  retrieving local climate zones patterns in satellite data.
  \textit{Environmental Challenges}, \textbf{5}, 100\,359,
  \doi{10.1016/j.envc.2021.100359}.

\bibitem[{Stevens et~al.(2020)}]{stevens2020sugar}
Stevens, B., and Coauthors, 2020: Sugar, gravel, fish and flowers: Mesoscale
  cloud patterns in the trade winds. \textit{Quarterly Journal of the Royal
  Meteorological Society}, \textbf{146~(726)}, 141--152.

\bibitem[{Strommen et~al.(2021)Strommen, Chantry, Dorrington,, and
  Otter}]{strommen2021topological}
Strommen, K., M.~Chantry, J.~Dorrington, and N.~Otter, 2021: A topological
  perspective on weather regimes. \textit{arXiv preprint arXiv:2104.03196},
  \eprint{2104.03196}.

\bibitem[{Sun et~al.(2021)Sun, Manchester,, and Chen}]{sun2021improved}
Sun, H., W.~Manchester, and Y.~Chen, 2021: Improved and interpretable solar
  flare predictions with spatial and topological features of the polarity
  inversion line masked magnetograms. \textit{Space Weather}, \textbf{19~(12)},
  \doi{10.1029/2021sw002837}.

\bibitem[{Topaz et~al.(2015)Topaz, Ziegelmeier,, and
  Halverson}]{topaz2015topological}
Topaz, C.~M., L.~Ziegelmeier, and T.~Halverson, 2015: Topological data analysis
  of biological aggregation models. \textit{PloS one}, \textbf{10~(5)},
  e0126\,383.

\bibitem[{Tymochko et~al.(2020)Tymochko, Munch, Dunion, Corbosiero,, and
  Torn}]{tymochko2020using}
Tymochko, S., E.~Munch, J.~Dunion, K.~Corbosiero, and R.~Torn, 2020: Using
  persistent homology to quantify a diurnal cycle in hurricanes.
  \textit{Pattern Recognition Letters}, \textbf{133}, 137--143.

\bibitem[{Xu et~al.(2021)Xu, Zhou, Fu, Zhou,, and Li}]{xu2021survey}
Xu, J., W.~Zhou, Z.~Fu, H.~Zhou, and L.~Li, 2021: A survey on green deep
  learning. \textit{arXiv preprint arXiv:2111.05193}.

\bibitem[{Yuval and O'Gorman(2020)Yuval, and O'Gorman}]{yuval2020stable}
Yuval, J., and P.~A. O'Gorman, 2020: Stable machine-learning parameterization
  of subgrid processes for climate modeling at a range of resolutions.
  \textit{Nature Communications}, \textbf{11~(1)},
  \doi{10.1038/s41467-020-17142-3}.

\bibitem[{Zhu et~al.(2017)Zhu, Tuia, Mou, Xia, Zhang, Xu,, and
  Fraundorfer}]{zhu2017deep}
Zhu, X.~X., D.~Tuia, L.~Mou, G.-S. Xia, L.~Zhang, F.~Xu, and F.~Fraundorfer,
  2017: Deep learning in remote sensing: A comprehensive review and list of
  resources. \textit{IEEE Geoscience and Remote Sensing Magazine},
  \textbf{5~(4)}, 8--36.

\end{thebibliography}


\end{document}